%% file: main.tex
\pdfoutput=1
\documentclass{article}
\usepackage{iclr2027_conference,times}

\input{math_commands.tex}

\input{imports}

\title{Semantically Steerable Image Tokenization \\ \\}

\author{Danilo de Goede$^{1,2}$, Mohammad Mahdi Derakhshani$^{2}$, Yuki M. Asano$^{3}$, Cees G. M. Snoek$^{1,2}$ \\
$^{1}$QUVA Lab, University of Amsterdam \\
$^{2}$VIS Lab, University of Amsterdam \\
$^{3}$Fundamental AI Lab, University of Technology Nuremberg \\
\texttt{\{d.degoede,m.m.derakhshani,c.g.m.snoek\}@uva.nl, yuki.asano@utn.de}
}

\iclrfinalcopy
\begin{document}

\maketitle
\lhead{Preprint}

\begin{abstract}
\input{sections/00_abstract}
\end{abstract}

\input{sections/01_introduction}
\input{sections/02_related_work}
\input{sections/03_method}
\input{sections/04_experiments}
\input{sections/05_conclusion}

\input{sections/ai_use_statement}

\subsubsection*{Acknowledgments}
This work is financially supported by Qualcomm Technologies Inc., the University of Amsterdam and the allowance Top consortia for Knowledge and Innovation (TKIs) from the Netherlands Ministry of Economic Affairs and Climate Policy. Cees G. M. Snoek is (partially) funded by the Horizon Europe project ELLIOT (GA No. 101214398). We thank SURF for the support in using the National Supercomputer Snellius.

\bibliography{references}
\bibliographystyle{iclr2027_conference}

\appendix
\crefalias{section}{appendix}
\crefalias{subsection}{appendix}
\raggedbottom
\clearpage

\section*{\LARGE Appendix}
\section*{Table of Contents}
\startcontents[appendices]
\printcontents[appendices]{l}{1}{\setcounter{tocdepth}{2}}

\input{sections/appendix/training_stages}
\input{sections/appendix/hyperparameters}
\input{sections/appendix/reproduction_report}
\input{sections/appendix/comparison_tokenizers}
\input{sections/appendix/tokens_vs_quality_metrics}
\input{sections/appendix/coco_evaluation}
\input{sections/appendix/token_budget_sweep_metrics}
\input{sections/appendix/qualitative_samples}
\input{sections/appendix/sem_knob_qualitative}
\input{sections/appendix/knn_pooling}
\input{sections/appendix/knn_k_sweep}
\input{sections/appendix/attentive_probing}

\end{document}

%% file: math_commands.tex
\usepackage{amsmath,amsfonts,bm}

\def\eqref#1{equation~\ref{#1}}

\def\1{\bm{1}}

\DeclareMathAlphabet{\mathsfit}{\encodingdefault}{\sfdefault}{m}{sl}
\SetMathAlphabet{\mathsfit}{bold}{\encodingdefault}{\sfdefault}{bx}{n}

%% file: imports.tex
\usepackage[T1]{fontenc}
\usepackage{microtype}

\usepackage{amsmath}
\usepackage{amssymb}
\usepackage{amsthm}
\usepackage{bm}

\usepackage{wrapfig}
\usepackage{graphicx}
\usepackage{subcaption}
\usepackage{booktabs}
\usepackage{epigraph}
\usepackage{float}
\usepackage{xcolor}
\usepackage{colortbl}
\usepackage{makecell}
\usepackage{adjustbox}
\definecolor{gistrow}{rgb}{0.85,0.91,0.98}
\usepackage{xspace}
\usepackage{soul}

\usepackage{enumerate}
\usepackage[inline,shortlabels]{enumitem}

\usepackage{hyperref}
\usepackage[capitalize,noabbrev,nameinlink]{cleveref}

\usepackage{titletoc}

\hypersetup{
    colorlinks=true,
    linkcolor=blue,
    filecolor=blue,
    urlcolor=blue,
    citecolor=blue
}

\crefname{appendix}{Appendix}{Appendices}
\Crefname{appendix}{Appendix}{Appendices}

\newcommand{\oursnew}[1]{\textcolor{teal}{#1}}

\newcommand{\karldagger}{KARL$^\dagger$\xspace}
\newcommand{\ourmethod}{GIST\xspace}
\newcommand{\ourmethodone}{GIST-1\xspace}

\newcommand{\lrec}{\ensuremath{\epsilon_{\mathrm{rec}}}\xspace}
\newcommand{\lsem}{\ensuremath{\epsilon_{\mathrm{sem}}}\xspace}

\newcommand{\enc}{\ensuremath{\mathrm{Enc}}\xspace}
\newcommand{\dec}{\ensuremath{\mathrm{Dec}}\xspace}
\newcommand{\quant}{\ensuremath{\mathrm{Quant}}\xspace}

\newcommand{\lrecEmb}{\ensuremath{\bm{\epsilon}_{\mathrm{rec}}}\xspace}
\newcommand{\lsemEmb}{\ensuremath{\bm{\epsilon}_{\mathrm{sem}}}\xspace}

\newcommand{\zOneD}{\ensuremath{\bm{Z}_\mathrm{1D}}\xspace}
\newcommand{\xTwoD}{\ensuremath{\bm{X}_\mathrm{2D}}\xspace}

\newcommand{\maxTokens}{\ensuremath{T_\mathrm{max}}\xspace}
\newcommand{\minTokens}{\ensuremath{T_\mathrm{min}}\xspace}

\newcommand{\xTwoDHat}{\ensuremath{\hat{\bm{X}}_{\mathrm{2D}}}\xspace}
\newcommand{\xTwoDTilde}{\ensuremath{\tilde{\bm{X}}_{\mathrm{2D}}}\xspace}

%% file: sections/00_abstract.tex
Adaptive image tokenizers compress an image into a variable number of tokens matched to its complexity, and recent work makes this allocation controllable: conditioned on a target reconstruction error, the tokenizer itself decides how many tokens an image needs.
However, the fidelity measure guiding this allocation shapes which aspects of the image are preserved under compression.
The pixel-wise error used in prior work is blind to the spatial structure that carries an image's meaning, and under heavy compression such tokenizers are prone to abruptly discarding tokens that encode essential content.
We introduce \ourmethod (\textbf{G}uided \textbf{I}mage \textbf{S}emantic \textbf{T}okenizer), an adaptive tokenizer that can be steered toward a requested \emph{semantic} fidelity.
Its token allocation is jointly controlled by pixel-level and semantic target errors, with semantic error measured in the feature space of a frozen DINOv3 encoder.
This joint control enables \emph{semantic protection}: with the semantic target fixed at zero, the reconstruction target can be relaxed to discard pixel detail while preserving what the image depicts.
In a controlled comparison on ImageNet-100, \ourmethod outperforms its pixel-controlled counterpart KARL at every token count, with up to $30\%$ lower LPIPS and $37\%$ lower DreamSim.
On average, lowering its semantic target reduces LPIPS $2\times$ faster per token than lowering the reconstruction target, and up to $5\times$ in the low-token regime.
Semantic fidelity is thus an effective control signal for adaptive tokenization: it lets a tokenizer see, and so preserve, what an image depicts.

%% file: sections/01_introduction.tex
\vspace{-0.5em}
\noindent\makebox[\linewidth][r]{%
  \begin{minipage}{0.75\textwidth}
    \small\raggedleft
    ``Humans can recognize the gist of a novel image in a single glance,\\
    independent of its complexity.''\\[0.2em]
    --- \textit{\citet{oliva2006gist}}%
  \end{minipage}\hspace*{\leftmargini}}
\vspace{-1em}

\section{Introduction}
\label{sec:introduction}

The visual world is inherently heterogeneous: some scenes depict a single object against a plain background, while others capture complex environments dense with detail.
Yet the image tokenizers underpinning modern generative models~\citep{esser2021vqgan, rombach2022ldm, chang2022maskgit} are blind to this difference --- they allocate the same number of tokens to every image, regardless of its content.
This allocation is wasteful: tokenizers exist to compress images into compact latent sequences over which generative modeling becomes tractable, and the length of this sequence directly governs training and inference cost.

A growing line of work therefore makes the token count adaptive~\citep{duggal2024alit, yan2024elastictok, miwa2025onedpiece, bachmann2025flextok}.
These tokenizers can encode an image into a variable number of tokens, but leave the choice of token count to a trial-and-error search at inference.
Ideally, the user would instead specify how faithful the encoding should be and let the tokenizer itself determine the token count.
KARL by~\citet{duggal2025karl} realizes exactly this: conditioned on the image and a target reconstruction error, it predicts the required token count in a single forward pass and halts the remaining tokens.
However, what the tokenizer prioritizes preserving depends on how fidelity is measured.
KARL measures fidelity with a pixel-wise reconstruction error that treats pixels independently and is blind to the spatial structure and co-dependence that carry an image's meaning.
An encoding can therefore meet the pixel target yet fail to preserve what the image depicts.
For example, a small bird against an open sky can become an unrecognizable blob of similar colors yet contribute little to the overall pixel error.
Indeed, we experimentally observe that under heavy token compression, KARL is prone to abruptly halting tokens that encode essential image content.

We introduce \ourmethod (\textbf{G}uided \textbf{I}mage \textbf{S}emantic \textbf{T}okenizer), an adaptive image tokenizer that can be steered toward a requested \emph{semantic} fidelity.
Its token allocation is jointly controlled by a reconstruction target error $\lrec$ and a semantic target error $\lsem$, with semantic error measured in the feature space of a frozen DINOv3 encoder~\citep{simeoni2025dinov3}.
These targets allow pixel-level and semantic fidelity to be specified independently: lowering $\lrec$ demands tokens that reproduce the image's exact pixels, while lowering $\lsem$ demands tokens that preserve what the image depicts.

\begin{figure}[t]
    \centering
    \includegraphics[width=\textwidth]{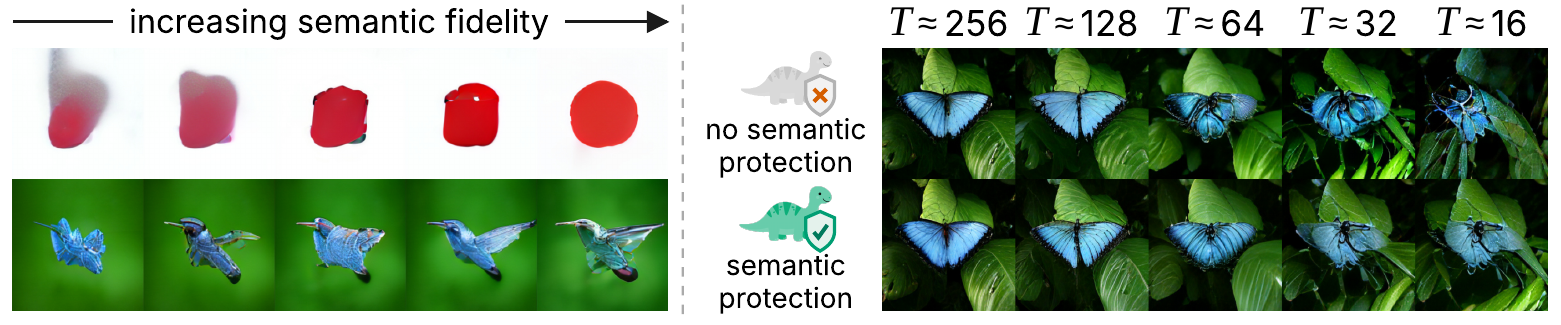}
    \caption{
        \textbf{Semantic control unlocks semantic protection.}
        (\emph{Left})~\ourmethod can be steered toward a requested semantic fidelity in DINOv3 feature space, as illustrated here with a red circle (top) and a hummingbird (bottom).
        Raising the requested fidelity steers the encoding from little more than the image's color distribution to a reconstruction that preserves what the image depicts.
        (\emph{Right})~Requesting full semantic fidelity turns this control into a guardrail: without it, compression into fewer tokens disintegrates the butterfly; with it, the subject remains recognizable down to 16 tokens.
    }
    \label{fig:gist_teaser}
\end{figure}

As illustrated in \cref{fig:gist_teaser}, this two-dimensional control unlocks a new capability we call \emph{semantic protection}: fixing the semantic target error at zero turns it into a guardrail under which the reconstruction target can be relaxed far more aggressively.
The encoder then discards tokens encoding redundant pixel detail while retaining those needed to preserve the image's semantic content, as captured in DINOv3 feature space.

We validate \ourmethod on ImageNet-100 against KARL, with all models trained from scratch under identical conditions.
\ourmethod Pareto-improves reconstruction quality at every number of active tokens, with up to $30\%$ lower LPIPS and $37\%$ lower DreamSim.
We further find that lowering the semantic target is more token-efficient than lowering the reconstruction target.
Per token, it reduces LPIPS $2\times$ faster on average, and up to $5\times$ in the low-token regime.

We find that these gains do not come from DINOv3 conditioning alone.
A variant with just the semantic target merely matches the baseline on reconstruction, yet its tokens are the most semantically informative of all models we train.
This suggests a division of labor.
Conditioning on DINOv3 feature error structures the latent space so that token complexity reflects the \emph{meaning} of an image rather than only its pixel-level appearance.
Semantic protection converts this structure into reconstruction gains.
The takeaway is simple: an adaptive tokenizer can only preserve what its fidelity measure can see, and a semantic measure lets it see what the image depicts.

%% file: sections/02_related_work.tex
\section{Related Work}
\label{sec:related_work}

\paragraph{Image Tokenization.}
Image tokenizers compress images into compact latent representations over which generative models can be efficiently trained.
Modeling a distribution directly in pixel space is difficult due to its high dimensionality, much of which encodes perceptually irrelevant detail.
Tokenizers address this by mapping images to a lower-dimensional latent space that retains only the information necessary for faithful reconstruction.
Variational Autoencoders (VAEs)~\citep{kingma2013vae} learn such representations as continuous latents and form the basis of latent diffusion models~\citep{rombach2022ldm}.
VQ-VAE~\citep{vandenoord2017vqvae,razavi2019vqvae2} and VQGAN~\citep{esser2021vqgan} instead apply vector quantization to produce discrete-valued tokens and have become the de facto tokenizers for masked~\citep{chang2022maskgit} and autoregressive~\citep{esser2021vqgan, tian2024autoregressive,sun2024arllama} image generation.
All of these tokenizers, however, share one structural constraint: the latent is a 2D grid whose size is fixed by the image resolution.

\paragraph{1D and Adaptive Tokenization.}
TiTok~\citep{yu2024titok} departs from the 2D grid: a ViT encoder distills the image into a short 1D sequence of register tokens~\citep{darcet2024vision}.
The sequence length, however, is still fixed at training time, independent of image content.
Subsequent work makes the length adaptive: ALIT~\citep{duggal2024alit} recurrently grows the 1D sequence, while ElasticTok~\citep{yan2024elastictok}, One-D-Piece~\citep{miwa2025onedpiece},  FlexTok~\citep{bachmann2025flextok}, and SMAP~\citep{li2026smap} train with tail-token dropout~\citep{rippel2014nesteddropout} so that any prefix of the ordered, coarse-to-fine token sequence remains decodable.
These methods make variable-length representations possible, but leave the choice of length to a search at inference: the user or the method itself must encode the image at several budgets and keep the shortest that suffices.

CAT~\citep{shen2025cat} sidesteps this search by scoring image complexity with an LLM and routing each image to one of three grid resolutions, yet the budget is set by an external predictor, at coarse granularity, and cannot be conditioned on a requested fidelity.
KARL~\citep{duggal2025karl} instead makes the tokenizer itself decide: conditioned on the image and a target reconstruction quality, it predicts the required token count in a single forward pass and halts the remaining tokens.
\ourmethod inherits KARL's single-pass, quality-conditioned design but extends the notion of quality itself: where KARL targets pixel-space reconstruction error alone, \ourmethod adds a second, semantic target, which we find improves token efficiency in the low-token regime.

\paragraph{Semantic Alignment of Tokenizers.}
A separate line of work distills the semantic knowledge of pretrained vision encoders into tokenizers.
An influential precursor is REPA~\citep{yu2025repa}, which aligns intermediate features of a diffusion transformer with DINOv2~\citep{oquab2024dinov2} features and accelerates convergence by more than an order of magnitude.
VA-VAE~\citep{yao2025vavae} moves the alignment from the generator into the tokenizer: a foundation-model alignment loss structures the VAE's latent space during training, which preserves reconstruction quality while making the latents far easier for diffusion transformers to model~\citep[see also][]{qu2026semtok}.
VFMTok~\citep{zheng2025vfmtok} and RAE~\citep{zheng2025rae} take this idea further by building the tokenizer directly on frozen foundation model features.
Adaptive-length tokenizers have adopted this alignment only in the decoder: FlexTok and Semanticist~\citep{wen2025semanticist} train their diffusion decoders with a REPA loss, and ReTok~\citep{fu2026retok} decays a decoder-side alignment loss along its ordered token sequence.

In all of these methods, however, semantic alignment is a fixed training-time bias: it shapes what the model encodes and improves output quality at a given token count, but plays no role in deciding how many tokens an image receives.
\ourmethod instead makes semantics an explicit control signal: the tokenizer is conditioned on a target quality in DINOv3~\citep{simeoni2025dinov3} feature space, so the token count is steered by the requested semantic fidelity rather than by pixel fidelity alone.
DINOv3 thus serves purely to \emph{measure} semantic fidelity for this conditioning --- its features never supervise the tokens themselves, unlike the alignment approaches above.

%% file: sections/03_method.tex
\section{Method}
\begin{figure}[t]
    \centering
    \includegraphics[width=\textwidth]{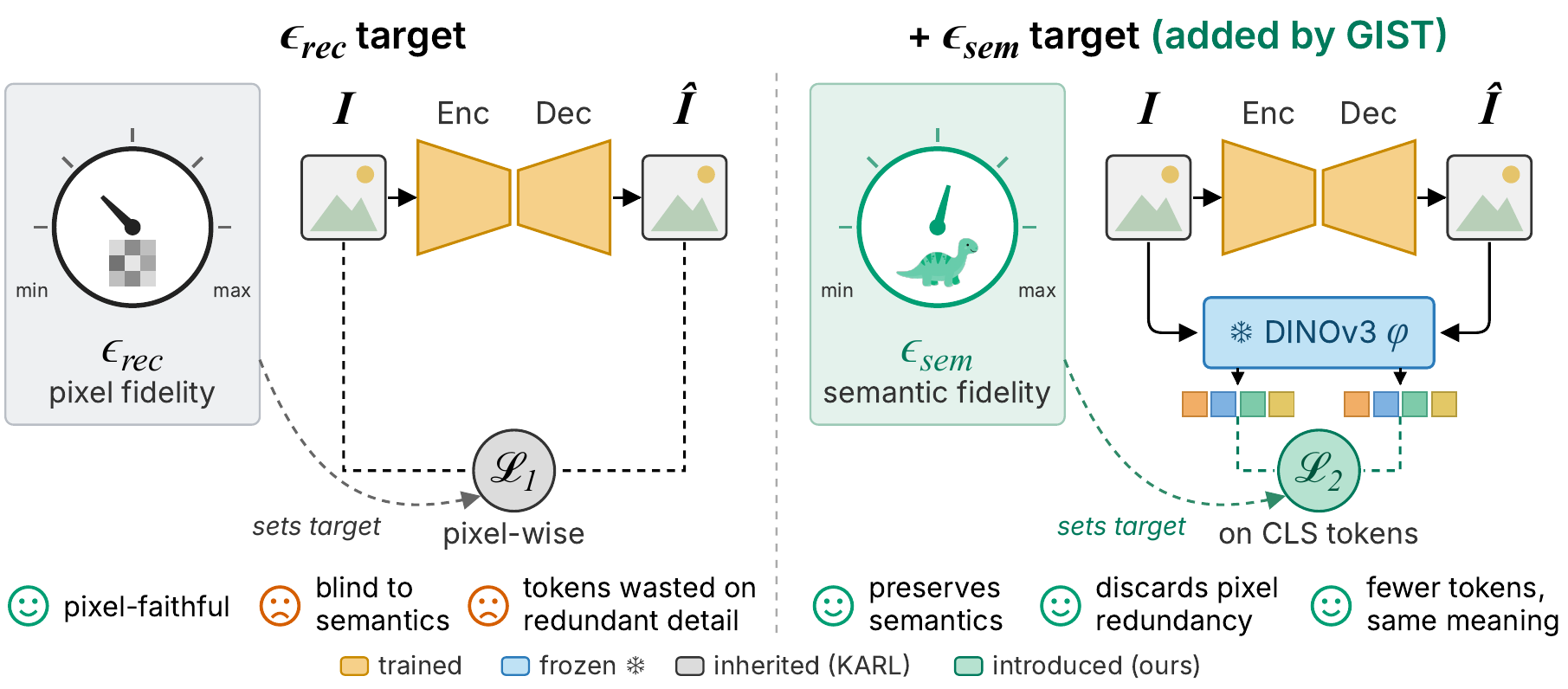}
    \caption{
        \textbf{\ourmethod makes semantic fidelity an explicit control signal.}
        (\emph{Left}) KARL~\citep{duggal2025karl} steers the tokenizer with a single reconstruction target $\smash{\lrec}$ for the pixel-wise $\smash{L_1}$ error between input $\smash{\bm{I}}$ and its reconstruction $\smash{\hat{\bm{I}}}$: encodings become pixel-faithful, but the error is blind to semantics, so tokens are prone to be wasted on redundant detail.
        (\emph{Right}) \ourmethod keeps $\smash{\lrec}$ and adds a second, semantic target $\smash{\lsem}$ for the error between DINOv3 CLS features of $\smash{\bm{I}}$ and $\smash{\hat{\bm{I}}}$: lowering $\smash{\lsem}$ demands tokens that preserve what the image depicts, so the encoder can discard redundant pixel detail.
    }
    \label{fig:gist_method}
\end{figure}

\ourmethod builds on KARL~\citep{duggal2025karl}, a single-pass adaptive 1D tokenizer that allocates a variable number of tokens per image based on a single reconstruction target error.
We introduce a second conditioning signal: a semantic target error, measured in the feature space of a frozen DINOv3 encoder~\citep{simeoni2025dinov3}.
This expands KARL's one-dimensional fidelity control into a two-dimensional space in which pixel-level and semantic fidelity can be targeted independently.
We illustrate the idea in \cref{fig:gist_method}.
We describe the architecture and the training procedure below, marking where \ourmethod deviates from KARL (\oursnew{in this color}).

\subsection{\ourmethod Architecture}
\label{sec:gist_architecture}
\ourmethod follows a standard encoder-decoder architecture that maps images to a 1D, variable-length latent space.
Given an input image, we first extract 2D patch embeddings $\xTwoD \in \mathbb{R}^{N \times D}$ from a pretrained VQGAN.
The encoder then distills these embeddings into learnable register tokens $\zOneD \in \mathbb{R}^{\maxTokens \times D}$~\citep{darcet2024vision}, following the 2D-to-1D token distillation of TiTok~\citep{yu2024titok} adopted by KARL.
The register tokens act as read-write storage during self-attention.

To control the complexity of the 1D tokenization, KARL conditions the encoder on a reconstruction \textbf{target error} $\lrec$, governing pixel-level fidelity.
We introduce a second, semantic target error $\lsem$, governing perceptual fidelity in the feature space of a frozen DINOv3 encoder~\citep{simeoni2025dinov3}.
Each target specifies an upper bound on its corresponding error: lowering it demands a more faithful encoding, and in turn more tokens.
Importantly, the two targets capture faithfulness of different kinds: lowering $\lrec$ steers the encoder toward tokens that reproduce the image's exact pixel values, while lowering $\lsem$ steers it toward tokens that preserve its semantic identity (what the image depicts, rather than the exact pixels used to depict it).
The targets thus control not only \emph{how many} tokens an image requires, but also \emph{what} those tokens encode.

We convert these continuous target errors into learnable conditioning signals by mapping them to the indices of their nearest valid upper-bound bins and embedding them via learnable dictionaries $\bm{E}_{\mathrm{rec}}$ and $\bm{E}_{\mathrm{sem}}$ to produce \emph{target embeddings}:
\begin{align}
    \lrecEmb &= \bm{E}_{\mathrm{rec}}[\mathrm{bin}(\lrec)], \\
    \oursnew{\lsemEmb} &= \oursnew{\bm{E}_{\mathrm{sem}}[\mathrm{bin}(\lsem)]}.
\end{align}
The encoder attends over the concatenated target embeddings, patch embeddings, and register tokens, allowing the registers to absorb information from all other tokens via self-attention.
We retain only the updated register outputs $\zOneD'$:
\begin{equation}
    \zOneD' = \enc\left([\lrecEmb; \oursnew{\lsemEmb}; \xTwoD; \zOneD]\right)_{\zOneD}.
\end{equation}
Depending on image complexity, only a subset of register tokens may be required to meet the target error bounds $(\lrec, \lsem)$.
To select the most suitable subset, a halting MLP~\citep{duggal2025karl} predicts a per-token halting probability from $\zOneD'$.
Tokens whose halting probability exceeds a predefined threshold (e.g., $\tau = 0.75$) are marked inactive and excluded from decoding.
\begin{align}
    \bm{\omega} &= \sigma\left(\text{MLP}_{\mathrm{halt}}(\zOneD')\right) \in [0, 1]^{\maxTokens}, \\
    \bm{m}_{\mathrm{halt}} &= \mathbb{I}(\bm{\omega}>\tau) \in \{0, 1\}^{\maxTokens}.
\end{align}
To prevent halted latents from interacting with the active sequence, we derive a joint self-attention mask $\bm{M}$ from $\bm{m}_{\mathrm{halt}}$ that blocks attention to and from any halted token. Under this mask, the decoder reconstructs the VQGAN patch embeddings by jointly attending over the quantized latent tokens $\smash{\hat{\bm{Z}}_{\mathrm{1D}} = \quant(\zOneD')}$ and a 2D grid of learnable query tokens $\smash{\xTwoDTilde}$:
\begin{equation}
    \xTwoDHat = \dec\left([\xTwoDTilde; \hat{\bm{Z}}_{\mathrm{1D}}], \text{mask}=\bm{M}\right)_{\xTwoDTilde}.
\end{equation}
At inference, tokenization requires only a single forward pass: the user specifies the target errors $\smash{(\lrec, \lsem)}$, and the encoder produces and halts tokens accordingly.
This requires neither iterative re-encoding nor a search over token counts.

\subsection{Dual Forward Pass Training}
\label{sec:gist_dual_forward}
The architecture of \ourmethod allows us to steer the complexity of the image encoding by conditioning on reconstruction and semantic target errors.
The values of these targets influence \emph{how} the encoder constructs suitable tokens $\zOneD'$ to meet the target error requirements, as well as \emph{how} the halting MLP processes these tokens to decide whether each of them should be halted or not.
This allows our method to dynamically infer both the appropriate number of tokens and the nature of each token for any given image, without resorting to trial-and-error or manual tuning.
\emph{How, then, can we train our method to be directly controllable by these targets?}

We achieve this by following~\citet{duggal2025karl} and optimizing \ourmethod using a dual forward pass based on the principles of upside-down reinforcement learning~\citep{schmidhuber2019upsidedownrl}.
During the \textbf{first forward pass}, we set $\lrec$ and $\lsem$ to $0$ to let \ourmethod aim for a perfect reconstruction under a uniformly sampled token budget $\smash{T \sim U(\minTokens, \maxTokens)}$.
We measure the error \ourmethod actually realizes in \emph{pixel space} by passing $\smash{\xTwoDHat}$ through the VQGAN decoder.
Whereas KARL measures the realized error solely through a pixel-wise reconstruction error, we additionally introduce a semantic error $\lsem^*$ that captures perceptual fidelity in the feature space of a frozen DINOv3 encoder:
\begin{align}
\lrec^* &= \tfrac{1}{3HW} \bigl\lVert \bm{I} - \mathrm{VQGAN}_{\mathrm{dec}}(\xTwoDHat) \bigr\rVert_1, \\
\oursnew{\lsem^*} &= \oursnew{\tfrac{1}{D_\phi} \bigl\lVert \phi(\bm{I}) - \phi\bigl(\mathrm{VQGAN}_{\mathrm{dec}}(\xTwoDHat)\bigr) \bigr\rVert_2^2},
\end{align}
where $\bm{I} \in [0,1]^{3 \times H \times W}$ is the input image and $\phi:\mathbb{R}^{3 \times H \times W} \to \mathbb{R}^{D_\phi}$ extracts the post-LayerNorm CLS token from the frozen DINOv3-L/16 image encoder.
These realized target errors $\epsilon^* = (\lrec^*, \lsem^*)$ serve as a task-specifying input reward for the second forward pass.
Intuitively, satisfying error bound  $\epsilon^*$ requires no more than $T$ tokens for the given image.

To leverage this insight, the \textbf{second forward pass} conditions the encoder on $\epsilon^*$ and expands the token budget by $\Delta T$ tokens, which the model is trained to disregard. Formally, the halting loss~\citep{duggal2025karl} is defined as:
\begin{equation}
\mathcal{L}_{\mathrm{halt}}(\boldsymbol{\omega})=\text{BCE}(\boldsymbol{\omega}_{0:T},\mathbf{0})+\text{BCE}(\boldsymbol{\omega}_{T:T+\Delta T},\mathbf{1}).
\end{equation}
This loss encourages the model to assign low halting probabilities to essential tokens ($\boldsymbol{\omega}_{0:T} \to \mathbf{0}$), and high halting probabilities to the redundant ones ($\boldsymbol{\omega}_{T:T+\Delta T} \to \mathbf{1}$).
By doing so, the model implicitly learns the relationship between the input image, the specified error bound $\epsilon^*$, and the requisite number of tokens $T$.

During both passes, we optimize \ourmethod using a single objective:
\begin{equation}
\mathcal{L}
= \mathcal{L}_{\mathrm{rec}}(\xTwoDHat, \xTwoD)
+ \beta \, \mathcal{L}_{\mathrm{quant}}(\hat{\bm{Z}}_{\mathrm{1D}})
+ \lambda \, \mathcal{L}_{\mathrm{halt}}(\bm{\omega}),
\label{eq:gist_loss_objective}
\end{equation}
where $\smash{\mathcal{L}_{\mathrm{rec}}(\xTwoDHat, \xTwoD)}$ is the reconstruction loss against the VQGAN tokenizer, $\mathcal{L}_{\mathrm{quant}}(\hat{\bm{Z}}_{\mathrm{1D}})$ is the standard VQ-VAE~\citep{vandenoord2017vqvae} commitment and codebook loss on the register outputs $\smash{\zOneD'}$, and $\smash{\mathcal{L}_{\mathrm{halt}}(\bm{\omega})}$ is the halting loss defined above. We set $\smash{\beta = \lambda = 1}$ throughout this paper.

\textbf{Note:}
The two passes share the main loss objective by construction: in the first pass we have $\Delta T = 0$, so the slice $\bm{\omega}_{T:T+\Delta T}$ is empty and $\mathcal{L}_{\mathrm{halt}}$ reduces to $\mathrm{BCE}(\bm{\omega}_{0:T}, \bm{0})$ to push all $T$ halting probabilities toward zero. In the second pass, both terms of $\mathcal{L}_{\mathrm{halt}}$ are active to teach the model to halt the $\Delta T$ redundant tokens while keeping the first $T$ tokens active.

\paragraph{Training Details.}
We train \ourmethod in two sequential stages: Latent Distillation Pre-Training and Pixel-Space Finetuning, both on ImageNet-100 (further details in \cref{app:app_training_stages}).
Full training requires approximately 470 H100 GPU-hours on 8 NVIDIA H100 GPUs ($\approx$2.5 days wall-clock).

%% file: sections/04_experiments.tex
\section{Experiments}
\label{sec:experiments}
In this section, we explore how modeling an image tokenizer's complexity along a semantic axis affects tokenizer behavior and performance.
We evaluate \ourmethod against KARL in terms of reconstruction quality, and visually compare how the reconstruction and semantic targets affect the image reconstructions for both methods.
Subsequently, we study the role of the semantic control in \ourmethod.

\subsection{Setup}

\paragraph{Model Variants.}
We consider two variants of our method:
\begin{enumerate*}[(i)]
    \item \ourmethod, which is conditioned on both a reconstruction and a semantic target error, and
    \item \ourmethodone, which is conditioned \textit{only} on a semantic target error.
\end{enumerate*}
\ourmethod decomposes complexity into two complementary axes and lets us control each independently, whereas \ourmethodone replaces the reconstruction target with the semantic one and lets us isolate their individual behavior.

\paragraph{Baselines.}
We evaluate \ourmethod against KARL as our primary baseline.
We note a slight performance discrepancy between the authors' published checkpoint and the model produced by their released training code.
To ensure a controlled comparison that isolates the effect of our contributions, all evaluated models are trained from scratch under identical conditions using the official, unmodified KARL training script.
We denote the resulting baseline as \karldagger. We document all reproduction details in \cref{app:karl_reproduction}, and situate \ourmethod among other tokenizers in \cref{app:comparison_tokenizers}.
Code, training configurations, and all trained checkpoints will be released upon publication.

\paragraph{Data and Evaluation.}

All models are trained on ImageNet-100~\citep{tian2020in100} and, unless stated otherwise, evaluated on its validation set.
Following KARL, we focus on per-image metrics.
Because the targets set per-image error bounds, the fidelity of each reconstruction to its own input is the quantity of interest.
Distribution-level metrics such as rFID~\citep{heusel2017gans} instead score whether reconstructions are realistic \textit{as a set}, and can reward a plausible image that is not a faithful reconstruction (see \cref{app:comparison_tokenizers}).
We provide training and evaluation details in \cref{app:training_details}.

\subsection{The Value of Semantic Steerability}
\begin{figure}[t]
    \centering
    \begin{minipage}[c]{0.766\textwidth}
        \centering
        \includegraphics[width=\linewidth]{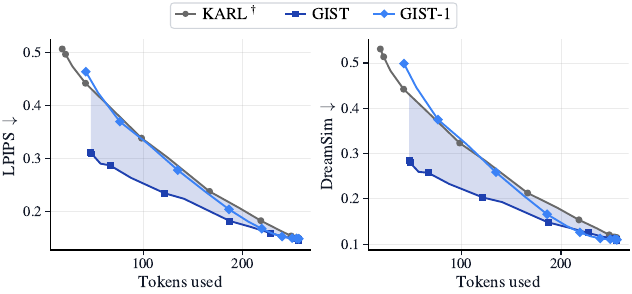}
    \end{minipage}\hfill
    \begin{minipage}[c]{0.201\textwidth}
        \raggedleft
        \scriptsize
        \setlength{\tabcolsep}{2pt}
        \begin{adjustbox}{max width=\linewidth}
        \begin{tabular}{l ccc}
        \toprule
        $\lrec$ & 0.05 & 0.10 & 0.16 \\
        \midrule
        \multicolumn{4}{l}{Tokens used $\downarrow$} \\
        \karldagger & \textbf{236} & \textbf{124} & \textbf{42} \\
        \rowcolor{gistrow}
        \ourmethod  & 242 & 141 & 67 \\
        \midrule
        \multicolumn{4}{l}{LPIPS $\downarrow$} \\
        \karldagger & 0.16 & 0.30 & 0.44 \\
        \rowcolor{gistrow}
        \ourmethod  & \textbf{0.15} & \textbf{0.22} & \textbf{0.29} \\
        \midrule
        \multicolumn{4}{l}{DreamSim $\downarrow$} \\
        \karldagger & 0.13 & 0.28 & 0.44 \\
        \rowcolor{gistrow}
        \ourmethod  & \textbf{0.12} & \textbf{0.19} & \textbf{0.26} \\
        \bottomrule
        \end{tabular}
        \end{adjustbox}
    \end{minipage}
    \caption{
        \textbf{Adding semantic control Pareto-improves reconstruction quality at every token count.}
        (\emph{Left}) Pareto fronts on the ImageNet-100 validation set, obtained by varying $\lrec$ for \karldagger and \ourmethod (fixing $\lsem = 0$), and by varying $\lsem$ for \ourmethodone.
        The shaded regions highlight \ourmethod's reduction in reconstruction error relative to \karldagger at matched token counts.
        (\emph{Right}) Tokens used, LPIPS, and DreamSim for \karldagger and \ourmethod at equal reconstruction targets, with the best values in bold.
        We report the same comparison on the full set of reconstruction quality metrics in \cref{app:app_tokens_vs_quality_metrics}.
    }
    \label{fig:tokens_vs_lpips_dreamsim_pareto}
\end{figure}

\paragraph{\ourmethod Pareto-improves reconstruction quality.}
In \cref{fig:tokens_vs_lpips_dreamsim_pareto}, we compare \ourmethod, \ourmethodone, and \karldagger along the token-fidelity Pareto front in LPIPS~\citep{zhang2018lpips} and DreamSim~\citep{fu2023dreamsim}, obtained by varying each method's target values.
We draw two conclusions.
First, \ourmethod achieves better reconstruction quality than \karldagger at every token count.
The gap widens as images are compressed into fewer tokens, with up to $30\%$ lower LPIPS and $37\%$ lower DreamSim.
This reconstruction advantage also extends to the multi-object scenes of MS COCO~\citep{lin2014coco}, without further training or adaptation (\cref{app:coco_evaluation}).
Second, \ourmethodone, which is conditioned only on the semantic target, performs on par with \karldagger.
This suggests that DINOv3 conditioning improves reconstruction quality only when complexity is disentangled into a reconstructive and a semantic axis.
We attribute the gap to the semantic target acting as a guardrail: fixing $\lsem = 0$ while relaxing $\lrec$ lets the encoder discard pixel-level detail while preserving the image's semantic identity.
We verify this in the remainder of our experiments.

In \cref{app:app_token_budget_sweep_metrics}, we repeat this comparison while varying the input token budget instead of the error targets.
We find that \ourmethod, \ourmethodone, and \karldagger perform on par in this setting, suggesting that the main benefit of our method comes from how conditioning on these targets organizes the latent space along complementary axes of complexity, rather than from base tokenizer capacity.

\paragraph{\ourmethod unlocks semantic protection.}
\begin{figure}[t]
    \centering
    \includegraphics[width=\textwidth]{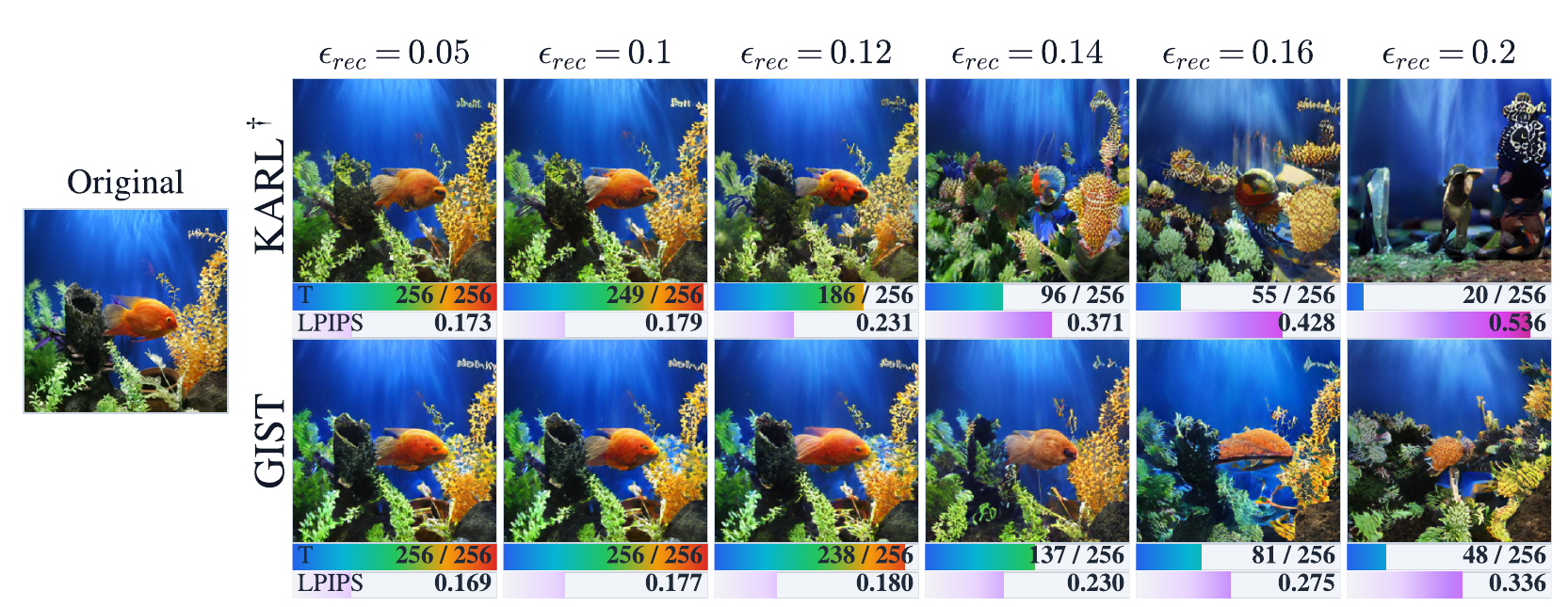}\\[-4pt]
    \includegraphics[width=\textwidth]{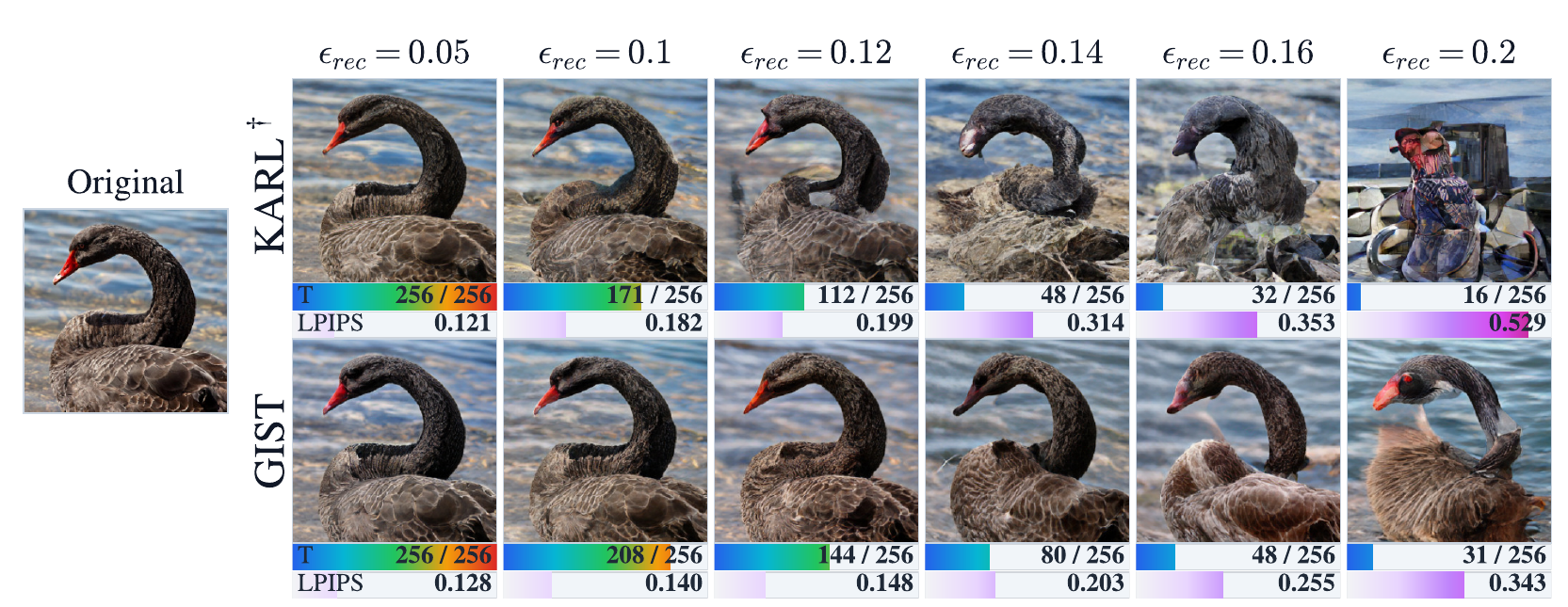}
    \caption{
        \textbf{Semantic protection keeps the subject recognizable under heavy token compression.}
        For each sample (\emph{Goldfish reef}, top; \emph{Black swan}, bottom), the upper row shows \karldagger reconstructions as we vary $\lrec$, and the lower row shows \ourmethod reconstructions with $\lsem$ fixed at $0$ and $\lrec$ varied over the same values.
        The far-left image is the original; token budget is fixed at $T = 256$; per-cell strips report the active token count and LPIPS.
        As more tokens are halted, \karldagger disintegrates the subject into artifacts that retain little beyond its color distribution, while \ourmethod keeps it recognizable.
        See \cref{app:app_qualitative_samples} for additional samples.
    }
    \label{fig:compare_karl_skarl_qualitative}
\end{figure}
To test the guardrail hypothesis visually, we qualitatively compare \karldagger and \ourmethod in \cref{fig:compare_karl_skarl_qualitative}.
As we raise the reconstruction target, we observe that \karldagger is prone to abruptly halting tokens that encode essential image content.
For \emph{Goldfish reef}, the fish becomes unrecognizable as soon as a substantial number of tokens are halted (here, $\lrec = 0.14$), and the image progressively collapses into incoherent artifacts that retain little of the original image beyond its color distribution.
We observe the same failure for \emph{Black swan}, where \karldagger preserves only the water background while the swan disintegrates into jigsaw-like fragments that retain only its color distribution.
\ourmethod, by contrast, is capable of \textit{reliably} halting more tokens before incurring such artifacts.
With $\lsem = 0$, raising $\lrec$ lets the encoder discard tokens encoding redundant pixel details while \textit{protecting} those indispensable to the image's semantic identity.
Across both samples and those in \cref{app:app_qualitative_samples}, this behavior is consistent --- the image's core gist survives more reliably even under heavy token compression.

\subsection{Understanding the Role of Semantic Control}
\begin{figure}[t]
    \centering
    \includegraphics[width=0.8\textwidth]{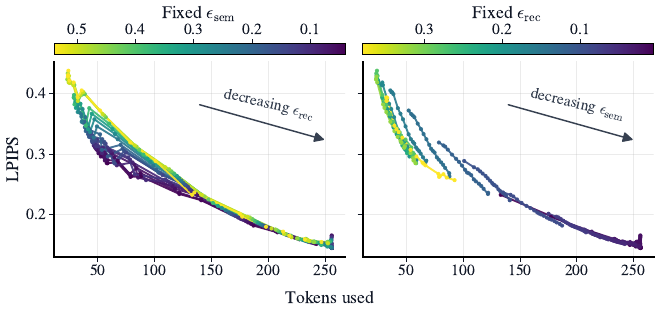}
    \caption{
        \textbf{Semantic control reduces LPIPS $2\times$ faster per token than pixel-level control.}
        LPIPS trajectories of \ourmethod as we vary one target with the other held fixed (shown by color).
        Lowering $\lrec$ (left) traces long, shallow curves that span nearly the full token range, whereas lowering $\lsem$ (right) drops LPIPS sharply over only a few added tokens.
    }
    \label{fig:iso_lpips}
    \label{fig:token_space_isocurves_lpips}
\end{figure}

\paragraph{Semantic control is more token-efficient.}
To better understand the role of each target in \ourmethod, we isolate their individual effects by holding one target ($\lrec$ or $\lsem$) fixed while gradually decreasing the other to trace the resulting LPIPS-token trajectories (\cref{fig:iso_lpips}).
Varying the two targets produces qualitatively different trajectories: lowering $\lrec$ (left) yields long, shallow curves that span nearly the full token range, whereas lowering $\lsem$ (right) yields steep, segmented curves that drop LPIPS sharply over only a few added tokens.
In other words, lowering the semantic target is more token-efficient at reducing LPIPS than lowering the reconstruction target.
On average, it reduces LPIPS $2\times$ faster per token, and up to $5\times$ faster in the low-token regime.

This efficiency is consistent with the semantic protection we observed in \cref{fig:compare_karl_skarl_qualitative}: we hypothesize that semantic control is more token-efficient because it prioritizes the tokens that carry the image's core gist.
The reconstruction target, by contrast, specifies a bound on pixel-wise $L_1$ error, which is blind to co-dependence and spatial structure that carry meaning in a visual signal.
Because this error provides no meaningful basis for prioritization, the encoder allocates tokens to faithfully reconstruct details that contribute little to perceptual quality.

\begin{figure}[t]
    \centering
    \begin{subfigure}{0.32\textwidth}
        \centering
        \parbox[c][1.70in][c]{\linewidth}{\centering
        \footnotesize
        \setlength{\tabcolsep}{4pt}
        \begin{tabular}{l ccc}
        \toprule
        & \multicolumn{3}{c}{Tokens} \\
        \cmidrule(lr){2-4}
        Method & 64 & 128 & 192 \\
        \midrule
        \multicolumn{4}{l}{kNN top-1 acc.\ (\%) $\uparrow$} \\
        \karldagger   & 21.9 & 23.3 & 24.2 \\
        \rowcolor{gistrow}
        \ourmethodone & \textbf{33.9} & \textbf{37.0} & \textbf{39.0} \\
        \rowcolor{gistrow}
        \ourmethod    & 28.0 & 28.4 & 28.6 \\
        \bottomrule
        \end{tabular}}
        \caption{\ourmethod vs. \karldagger vs. \ourmethodone}
        \label{fig:iso_knn_comparative}
    \end{subfigure}
    \hfill
    \begin{subfigure}{0.66\textwidth}
        \centering
        \includegraphics[width=\textwidth]{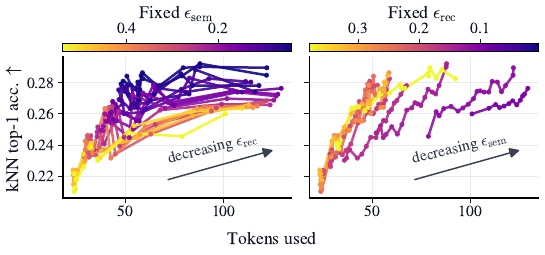}
        \caption{Varying one target at a time}
        \label{fig:iso_knn}
    \end{subfigure}
    \caption{
        \textbf{Semantic conditioning makes tokens more semantically informative.}
        (a)~kNN accuracy of \ourmethod, \karldagger, and \ourmethodone as we vary each method's target values (same setup as~\cref{fig:tokens_vs_lpips_dreamsim_pareto}), with accuracies linearly interpolated at matched mean active token counts.
        \ourmethod and \ourmethodone outperform \karldagger at every token count, and \ourmethodone performs best despite performing on par with \karldagger and worse than \ourmethod on reconstruction quality.
        (b)~kNN accuracy trajectories of \ourmethod as we vary one target with the other held fixed (shown by color).
        Lowering $\lsem$ increases kNN accuracy at a much steeper rate than lowering $\lrec$.
        Accuracies are computed by mean pooling over the active tokens; \cref{app:knn_pooling} reports the same comparison under max pooling.
    }
\end{figure}

\paragraph{Semantic conditioning makes tokens more semantically informative.}
Thus far, we evaluated \ourmethod through the quality of its reconstructions.
We now turn our attention to the latent tokens themselves and ask how conditioning on $\lsem$ affects their semantic informativeness.
In \cref{fig:iso_knn_comparative}, we compare the kNN classification accuracy of \ourmethod, \ourmethodone, and \karldagger as we vary each method's target values (protocol in \cref{app:evaluation_protocol}).
\ourmethod and \ourmethodone both achieve a higher kNN accuracy than \karldagger at any number of active tokens (+3--7 and +7--17 points, respectively, across 40--256 tokens).
Notably, \ourmethodone performs best of the three, despite performing on par with \karldagger and worse than \ourmethod on reconstruction quality (\cref{fig:tokens_vs_lpips_dreamsim_pareto}).
The same ordering holds under a trained attentive probe, with narrower but consistent gaps between methods (\cref{app:attentive_probing}).
We also visualize \ourmethod's kNN-token trajectories as we vary $\lrec$ and $\lsem$ in isolation to study their individual effect in \ourmethod's latent space in \cref{fig:iso_knn}.
Consistent with our earlier findings, lowering $\lsem$ increases kNN accuracy at a much steeper rate per added token than lowering $\lrec$.

Together, these results suggest that conditioning on $\lsem$ structures the latent space of \ourmethod and \ourmethodone so that token complexity reflects the \textit{meaning} of an image rather than only its pixel-level appearance.
However, this semantic structure can only be used to improve reconstruction quality when we introduce $\lsem$ as a second target, where it can act as a guardrail that protects the image's semantic identity while relaxing $\lrec$ allows the encoder to discard redundant pixel detail (\cref{fig:tokens_vs_lpips_dreamsim_pareto,fig:compare_karl_skarl_qualitative}).
In other words, semantic protection converts this structure into reconstruction gains.

%% file: sections/05_conclusion.tex
\section{Conclusion}
\label{sec:conclusion}
We have presented \ourmethod, an adaptive image tokenizer that can be steered toward a requested \emph{semantic} fidelity.
Its token allocation is jointly controlled by a pixel-wise reconstruction target and a semantic target in DINOv3 feature space, allowing pixel-level and semantic fidelity to be specified independently.
Our experiments point to a division of labor between the two roles the semantic target plays.
During training, the DINOv3 target structures the latent space so that token complexity reflects the meaning of an image rather than only its pixel-level appearance.
At inference, it acts as a guardrail: with $\lsem$ fixed at zero, the reconstruction target can be relaxed freely while the encoder protects the tokens indispensable to the image's semantic identity.
Under this guardrail, \ourmethod Pareto-improves reconstruction quality over \karldagger at every token count, and its tokens themselves become more semantically informative.

Our work also has its limitations.
First, our evaluation is deliberately controlled rather than competitive: all models are trained from scratch on ImageNet-100 over a pretrained VQGAN base.
Accordingly, we claim no state of the art.
The evidence supports the narrower point that semantic fidelity is an effective control signal for adaptive tokenization.
Second, the same principle that motivates \ourmethod also bounds it: a tokenizer can only preserve what its fidelity measure can see, and we instantiate that measure with a single global feature, the DINOv3 CLS token.
For images whose meaning is distributed across many regions, such as a document page or a busy scene in which several objects matter equally, one global summary may under-specify what to protect.
Patch-level semantic targets are a natural extension.

More broadly, we see the DINOv3 target as one instance of a general recipe: any measure of what matters can serve as a conditioning target.
Because the measure only conditions the tokenizer and is never differentiated through, it need not even be differentiable.
A different vision encoder, a text encoder, or a task-specific score such as detection or OCR accuracy could take its place.
In this way, \ourmethod can be tailored to the kind of information its tokens should preserve.
The tokens themselves also point beyond reconstruction.
Semantic alignment is known to ease generative training~\citep{yu2025repa, yao2025vavae}, and we suspect \ourmethod's tokens offer a favorable interface for generative models trained on top of them.
Whether semantic steerability carries over to generation is a question we leave open for future work.

Humans recognize the gist of a novel image in a single glance, independent of its complexity; we hope \ourmethod is a step toward tokenizers that see images the same way.

%% file: sections/ai_use_statement.tex
\subsection*{AI use statement}
In this work, we used generative AI tools (Anthropic's Claude, including the Claude Code coding assistant) for the following tasks in the required-disclosure category: implementing methods, in the form of software supporting our experiments (evaluation and job scripts, and experiment execution and management on our compute cluster); interpreting results, in the form of computing the aggregate numerical values reported in the text (e.g., averages over per-setting metrics) programmatically from the experiment output files; and designing or providing feedback on research methodology or experiments, in the form of brainstorming individual design choices (e.g., suggesting isocurves as the visualization of the four-dimensional relation between the two target qualities and the resulting token counts and reconstruction quality, and discussing which attentive-probe design to use), with the final choices made by us.

We have not used generative AI tools to help develop theoretical models or conceptual frameworks, to propose or refine hypotheses, or to generate synthetic datasets; and the rest of the required disclosure tasks (formulating mathematical claims, providing critical ingredients for proving mathematical claims, assisting in the writing of proofs, assisting with translation, cleaning and reformatting datasets, and supporting qualitative and thematic data analysis) are not applicable to this work.

Additionally, we used generative AI tools for tasks in the recommended-disclosure category: drafting and iteratively revising parts of the paper to improve clarity and readability; proposing candidate titles (the final title is our own); designing the introduction and method-overview figures, and creating the scripts that generate our plots and result tables; creating and editing software code; suggesting experimental parameters; summarizing or analyzing existing literature, in the form of answering targeted questions about the content of individual papers; identifying relevant literature, in the form of surfacing a few additional related-work references; and reformatting bibliography entries, in the form of removing redundant fields from entries we collected from DBLP (generative AI was not used to generate the citation entries themselves).

We have reviewed all AI-assisted work: we read, revised, and approved every AI-drafted sentence; we verified and tested AI-generated code for correctness; all AI-computed numbers and tables were derived programmatically from the experiment result files (never produced from model memory) and we verified them against those files; and we manually inspected AI edits to the bibliography line by line.
We take responsibility for the final content of this work, including text, claims or artifacts produced with the aid of generative AI.

%% file: sections/appendix/training_stages.tex
\clearpage
\section{Training and Evaluation Details}
\label{app:training_details}
This section documents the complete experimental setup: the two training stages (\cref{app:app_training_stages}), every hyperparameter of the models and of both stages (\cref{app:hyperparameters}), and the evaluation protocol (\cref{app:evaluation_protocol}).

\subsection{Two-Stage Training}
\label{app:app_training_stages}

We train \ourmethod in two sequential training stages that share the overall objective, but differ in the space in which $\mathcal{L}_{\mathrm{rec}}$ is evaluated.

\begin{itemize}
    \item \textbf{Stage 1 -- Latent Distillation Pre-Training:}
    We distill a pretrained, frozen VQGAN~\citep{esser2021vqgan} into the variable-length 1D \ourmethod encoder.
    The \ourmethod decoder predicts a VQGAN codebook distribution at every 2D grid position.
    Subsequently, $\mathcal{L}_{\mathrm{rec}}$ is computed directly in the VQGAN token space using a cross-entropy over codebook indices and an MSE between predicted and ground-truth code embeddings.
    This avoids the costly pixel decoding during this pre-training stage.

    \item \textbf{Stage 2 -- Pixel-Space Finetuning:}
    We finetune \ourmethod jointly with the VQGAN to reconstruct high-fidelity images in pixel space.
    To achieve this, we pass the output of the \ourmethod decoder into the VQGAN decoder to reconstruct the original image in pixel space.
    During this stage, $\mathcal{L}_{\mathrm{rec}}$ operates in pixel space and comprises a pixel-wise $\ell_1$ term, an LPIPS~\citep{zhang2018lpips} perceptual term, and an adversarial loss against a learned discriminator~\citep{esser2021vqgan}.
\end{itemize}

%% file: sections/appendix/hyperparameters.tex
\subsection{Hyperparameters}
\label{app:hyperparameters}

All models in this paper (\karldagger, \ourmethod, and \ourmethodone) are trained with the released KARL training script~\citep{duggal2025karl}, which we leave unchanged except for the components that define our method: semantic conditioning and the frozen DINOv3 encoder~\citep{simeoni2025dinov3} that measures the semantic error.
\Cref{tab:hyperparameters} lists the architecture, data, and compute configuration of every model, and \cref{tab:optimization} the optimization settings of the two training stages of \cref{app:app_training_stages}.

\newcolumntype{Y}{>{\centering\arraybackslash}p{0.17\linewidth}}

\paragraph{Token budgets during training.}
In the first forward pass of \cref{sec:gist_dual_forward}, we draw the token budget $T$ uniformly at random, with $\minTokens = 16$ and $\maxTokens = 256$.
The second forward pass always uses a budget of $256$ tokens, i.e., $\Delta T = \max(0, 256 - T)$, and conditions the encoder on the realized errors $\epsilon^*$ after mapping each of them to the nearest bin that upper-bounds it.

\subsection{Evaluation Protocol}
\label{app:evaluation_protocol}
\paragraph{Reconstruction.}
We compute all reconstruction metrics on the 5{,}000 images of the ImageNet-100 validation set at a $256 \times 256$ center crop, using the EMA weights of the final stage-2 checkpoint.
In the token-budget comparisons (\cref{app:comparison_tokenizers,app:app_token_budget_sweep_metrics}), we evaluate the budgets $T \in \{32, 64, 96, 128, 160, 192, 224, 256\}$ with both targets fixed at $0$.
In the target-error comparisons (\cref{fig:tokens_vs_lpips_dreamsim_pareto,fig:token_space_isocurves_lpips}), we fix $T = 256$ and vary each target over every bin of its dictionary: $\lrec$ for \karldagger, the full $(\lrec, \lsem)$ grid for \ourmethod, and $\lsem$ for \ourmethodone.
We count a token as active if its halting probability satisfies $\omega \leq \tau = 0.75$, and report the average number of active tokens over the validation set.
We use linear interpolation between evaluated points to compute reported gains at matched mean active token counts.
We use the AlexNet variant of LPIPS~\citep{zhang2018lpips}, the default DreamSim ensemble~\citep{fu2023dreamsim}, and SSIM~\citep{wang2004ssim} and PSNR on $[0, 1]$-scaled images.

\paragraph{kNN classification.}
As in \cref{app:knn_pooling}, we represent each image by the mean (or max) over its active register tokens of the 1024-d encoder outputs $\zOneD'$ after the final LayerNorm, i.e., before the projection to the 12-d quantized latents.
We use these features for both kNN classification and attentive probing to assess how semantic conditioning shapes the encoder representation, consistent with the encoder-level evaluation in ALIT~\citep{duggal2024alit}.
These evaluations measure the accessibility of class information before projection and quantization but do not directly assess its preservation in the discrete token sequence.

We $\ell_2$-normalize these features and classify each query by an unweighted majority vote over its $k = 20$ nearest neighbors under cosine similarity, using the ImageNet-100 training set as the gallery and the validation set as queries, both at a $256 \times 256$ center crop and with $T = 256$.
\Cref{app:knn_k_sweep} repeats the comparison for $k \in \{10, 20, 100, 200\}$ and shows the ordering is unchanged.

\paragraph{Attentive probing.}
The probe of \cref{app:attentive_probing}, adopted from V-JEPA~\citep{bardes2024vjepa}, consists of a single learnable query that cross-attends to the frozen 1024-d encoder features used for the kNN evaluation (16 heads, MLP ratio 4, pre-LN), followed by a linear classifier over the 100 classes.
We mask halted tokens ($\omega > 0.75$) out of this cross-attention, so the probe attends only over active tokens.
For every target configuration and token budget, we train a separate probe for 10 epochs on the ImageNet-100 training set with AdamW (learning rate $10^{-3}$, weight decay $0.01$, batch size 64, cosine decay to 0 without warmup), a cross-entropy loss, and no data augmentation beyond the center crop.
We report the best and last validation top-1 accuracy over the 10 epochs.

\begin{table}[H]
    \caption{\textbf{Model, data, and compute configuration of all models trained in this paper.}
    The cells spanning all columns apply to every model.
    The models differ only in which target errors they are conditioned on and in the frozen DINOv3 encoder of \ourmethod and \ourmethodone.}
    \label{tab:hyperparameters}
    \centering
    \begin{adjustbox}{max width=\linewidth}
    \begin{tabular}{@{}l|YYY@{}}
    \toprule
    Configuration & \karldagger & \ourmethod & \ourmethodone \\

    \midrule
    Encoder / decoder depth & \multicolumn{3}{c}{8 / 8} \\
    Encoder / decoder width & \multicolumn{3}{c}{1024 / 1024} \\
    Attention heads & \multicolumn{3}{c}{8} \\
    MLP ratio & \multicolumn{3}{c}{4} \\
    Feedforward activation & \multicolumn{3}{c}{GELU} \\
    Trainable parameters & \multicolumn{3}{c}{216M} \\
    Image resolution & \multicolumn{3}{c}{$256^2$} \\
    Patch size & \multicolumn{3}{c}{$16 \times 16$ (256 patch tokens)} \\
    Patch embedding & \multicolumn{3}{c}{\makecell{Convolutional patch features (768-d), concatenated\\ with VQGAN encoder features (256-d)}} \\
    Halting MLP & \multicolumn{3}{c}{Linear(1024, 512) -- Tanh -- Linear(512, 1)} \\
    Halting threshold $\tau$ & \multicolumn{3}{c}{0.75} \\
    1D codebook & \multicolumn{3}{c}{$4096 \times 12$-d, $\ell_2$-normalized} \\
    Commitment cost & \multicolumn{3}{c}{0.25} \\

    \midrule
    Base 2D tokenizer & \multicolumn{3}{c}{VQGAN~\citep{esser2021vqgan}} \\
    VQGAN checkpoint & \multicolumn{3}{c}{MAGE ImageNet-1k release~\citep{li2023mage}} \\
    VQGAN downsampling factor & \multicolumn{3}{c}{16} \\
    VQGAN codebook & \multicolumn{3}{c}{$1024 \times 256$-d} \\
    VQGAN parameters & \multicolumn{3}{c}{54.5M} \\

    \midrule
    Reconstruction embedding $\bm{E}_{\mathrm{rec}}$ & \checkmark & \checkmark & --- \\
    Semantic embedding $\bm{E}_{\mathrm{sem}}$ & --- & \checkmark & \checkmark \\
    Semantic encoder $\phi$ & --- & \multicolumn{2}{c}{\makecell{DINOv3 ViT-L/16\\ \citep{simeoni2025dinov3}}} \\
    Semantic encoder parameters & --- & \multicolumn{2}{c}{303M (frozen)} \\
    Semantic feature & --- & \multicolumn{2}{c}{Post-LN CLS token (1024-d)} \\

    \midrule
    Dataset & \multicolumn{3}{c}{ImageNet-100~\citep{tian2020in100}} \\
    Training / evaluation images & \multicolumn{3}{c}{$\approx$127k / 5{,}000} \\
    Augmentations & \multicolumn{3}{c}{\makecell{\texttt{RandomResizedCrop} (scale $[0.8, 1.0]$),\\ \texttt{RandomHorizontalFlip}}} \\
    Evaluation preprocessing & \multicolumn{3}{c}{$256 \times 256$ center crop} \\

    \midrule
    GPUs & \multicolumn{3}{c}{$8 \times$ NVIDIA H100 80GB} \\
    Training cost & \multicolumn{3}{c}{$\approx$470 GPU-hours ($\approx$2.5 days) per model} \\

    \bottomrule
    \end{tabular}
    \end{adjustbox}
\end{table}

\begin{table}[p]
    \caption{\textbf{Optimization settings of the two training stages.}
    Both stages use the released KARL recipe and are identical for \karldagger, \ourmethod, and \ourmethodone.
    Effective learning rates follow $\mathrm{lr} = \mathrm{base\ lr} \times \mathrm{batch\ size} / 256$.
    Loss weights $\beta$ and $\lambda$ refer to the objective in \cref{eq:gist_loss_objective}.}
    \label{tab:optimization}
    \centering
    \begin{adjustbox}{max width=\linewidth}
    \begin{tabular}{@{}l|cc@{}}
    \toprule
    Configuration & Stage 1: latent distillation & Stage 2: pixel-space finetuning \\

    \midrule
    Initialization & Random & Stage-1 EMA weights \\
    Epochs & 200 & 400 \\
    Warmup epochs & \multicolumn{2}{c}{20} \\
    Learning rate schedule & \multicolumn{2}{c}{Linear warmup, cosine decay to 0} \\
    Optimizer & \multicolumn{2}{c}{AdamW} \\
    Optimizer momentum & \multicolumn{2}{c}{$\beta_1, \beta_2 = 0.9, 0.95$} \\
    Base learning rate & \multicolumn{2}{c}{$1 \times 10^{-4}$} \\
    Effective learning rate & $2 \times 10^{-4}$ & $5.6 \times 10^{-5}$ \\
    Batch size & 512 (64 per GPU) & 144 (18 per GPU) \\
    Weight decay & \multicolumn{2}{c}{0.05} \\
    Gradient clipping norm & \multicolumn{2}{c}{3.0} \\
    Model EMA decay & \multicolumn{2}{c}{0.9999} \\
    Data type & \multicolumn{2}{c}{\texttt{float16} autocast} \\

    \midrule
    Reconstruction loss $\mathcal{L}_{\mathrm{rec}}$ & \makecell{Cross-entropy on VQGAN code indices\\ $+$ MSE to VQGAN code embeddings} & \makecell{$L_1$ $+$ LPIPS-VGG~\citep{zhang2018lpips}\\ $+$ hinge adversarial loss} \\
    Loss weights $\beta$, $\lambda$ & \multicolumn{2}{c}{1, 1} \\
    Adversarial loss weight & --- & \makecell{Adaptive~\citep{esser2021vqgan}, scaled by\\ 0 (epochs 0--20), 0.2 (21--100), 0.8 (101--399)} \\
    Discriminator & --- & PatchGAN (3 layers, 64 base channels) \\
    Number of discriminators & --- & 8 (one per 32-token bucket of active tokens) \\
    Discriminator optimizer & --- & AdamW, same learning rate \\
    Discriminator update & --- & Every 9th iteration \\
    VQGAN & Frozen & Finetuned (encoder and decoder) \\

    \bottomrule
    \end{tabular}
    \end{adjustbox}
\end{table}

%% file: sections/appendix/reproduction_report.tex
\clearpage
\section{Reproduction of the KARL Baseline}
\label{app:karl_reproduction}

In this paper, our primary baseline is \karldagger, which denotes KARL retrained with its official, unmodified training code and the arguments of its released run scripts.
We observe a small discrepancy between KARL's published numbers~\citep{duggal2025karl} and those we obtain by retraining the model with its unmodified training script.
The discrepancy can originate on either
\begin{enumerate*}[(i)]
    \item the \textit{evaluation side}, if our protocol differs from the authors', or
    \item the \textit{training side}, if the retrained model differs from the published checkpoint through code, configuration, hardware, or seed.
\end{enumerate*}
To isolate the source of the discrepancy, we compare three configurations, all run through the same evaluation pipeline:
\begin{itemize}
    \item \textbf{KARL (published).} The numbers reported in Table 1 of~\citet{duggal2025karl}.
    \item \textbf{KARL (public ckpt., our eval).} The official KARL checkpoint published on GitHub (\texttt{karl\_small\_vqgan\_quantized\_latents.pth}, trained on ImageNet-100).
    \item \textbf{\karldagger\ (reproduction, our eval).} KARL retrained from the official, unmodified training script.
    We report three independent retrains to separate the training-side gap from seed variance.
\end{itemize}
We train on 8 H100 GPUs (two nodes of four) in the environment specified by the repository's \texttt{environment.yaml} (PyTorch 2.3.0, CUDA 12.1), with 8 processes, per-GPU batch sizes of 64 and 18, and 200 and 400 epochs for the latent-distillation and finetuning stages, respectively.

We summarize the results in \cref{tab:karl_reproduction}, and draw three conclusions.
\begin{itemize}
    \item \textbf{Our evaluation protocol reproduces the published numbers.} Row~2 matches row~1 to within $0.01$ on every metric at every reported budget.
    Our evaluation protocol is thus not the source of the discrepancy.

    \item \textbf{The gap is on the training side, exceeds seed variance, and is concentrated in pixel-level metrics.}
    \karldagger (rows~3--5) performs slightly worse than the released checkpoint (row~2) on L1 Loss $\times$10 ($+0.04$ to $+0.10$) and SSIM ($-0.02$ to $-0.04$) at every budget; on LPIPS and DreamSim the gap is $+0.03$ at 32 tokens, at most $+0.02$ at 64 tokens, and within $\pm 0.01$ from 128 tokens on.
    Across the three retrains, the spread between seeds is $\leq 0.02$ on every metric and budget.
    This spread is far smaller than the gap to the published checkpoint, so the discrepancy is systematic rather than an artifact of seed variance.
    Its cause cannot be determined from the released artifacts: the published checkpoint stores only EMA weights and carries no record of the arguments, epoch count, or optimizer state that produced it.

    \item \textbf{The gap cancels in all controlled comparisons made in this paper.}
    Because \karldagger, \ourmethod, and \ourmethodone share the same training and evaluation protocol, the gap introduced by the training script affects each method equally.
    The \ourmethod and \ourmethodone rows make this concrete: both match the \karldagger mean to within $0.01$ on every metric and budget, i.e., within seed spread, and sit below the published checkpoint by the same systematic gap.
    Our claims concern how the targets guide token allocation (\cref{sec:experiments}), not base capacity, and are made relative to \karldagger at matched base capacity.
    Comparing \ourmethod against the published checkpoint would instead confound our contribution with an unexplained training difference.
    We therefore adopt \karldagger as our controlled baseline.
\end{itemize}

\begin{table}[H]
\centering

\caption{\textbf{Reproduction of the KARL baseline.}
We compare KARL's published numbers~\citep{duggal2025karl} against the official checkpoint and \karldagger (retrained from the official, unmodified code).
Our evaluation of the official checkpoint (row 2) reproduces the published numbers (row 1) to within $0.01$.
The retrained models (rows 3--5) perform slightly worse than the official checkpoint, mainly on L1 and SSIM.
Every std is at most $0.009$, well below this gap, so the gap is systematic rather than an artifact of seed variance.
The last two rows report \ourmethod and \ourmethodone; both match \karldagger.
Every model trained from the released code thus shares the same base capacity, below the official checkpoint by the same gap.}
\label{tab:karl_reproduction}
\small
\setlength{\tabcolsep}{3.5pt}
\resizebox{\textwidth}{!}{%
\begin{tabular}{l cccc cccc cccc cccc}
\toprule
& \multicolumn{4}{c}{LPIPS $\downarrow$}
& \multicolumn{4}{c}{DreamSim $\downarrow$}
& \multicolumn{4}{c}{L1 Loss$\times$10 $\downarrow$}
& \multicolumn{4}{c}{SSIM $\uparrow$} \\
\cmidrule(lr){2-5}\cmidrule(lr){6-9}\cmidrule(lr){10-13}\cmidrule(lr){14-17}
& 32 & 64 & 128 & 256 & 32 & 64 & 128 & 256
& 32 & 64 & 128 & 256 & 32 & 64 & 128 & 256 \\
\midrule
KARL (published)~\citep{duggal2025karl} & \textbf{0.30} & \textemdash & \textbf{0.19} & \textbf{0.14} & \textbf{0.28} & \textemdash & 0.16 & \textbf{0.11} & \textbf{1.13} & \textemdash & \textbf{0.80} & \textbf{0.70} & \textbf{0.34} & \textemdash & 0.44 & \textbf{0.49} \\
KARL (public ckpt., our eval)           & \textbf{0.30} & \textbf{0.24} & \textbf{0.19} & \textbf{0.14} & \textbf{0.28} & \textbf{0.21} & 0.16 & \textbf{0.11} & 1.14 & \textbf{0.98} & \textbf{0.80} & \textbf{0.70} & \textbf{0.34} & \textbf{0.38} & \textbf{0.45} & \textbf{0.49} \\
\midrule
\karldagger (reproduction, seed 1)         & 0.33 & 0.26 & \textbf{0.19} & 0.15 & 0.31 & 0.23 & \textbf{0.15} & \textbf{0.11} & 1.24 & 1.05 & 0.87 & 0.74 & 0.31 & 0.35 & 0.41 & 0.46 \\
\karldagger (reproduction, seed 2)         & 0.33 & 0.26 & \textbf{0.19} & 0.15 & 0.31 & 0.22 & \textbf{0.15} & \textbf{0.11} & 1.24 & 1.03 & 0.88 & 0.75 & 0.31 & 0.36 & 0.41 & 0.46 \\
\karldagger (reproduction, seed 3)         & 0.33 & 0.26 & \textbf{0.19} & \textbf{0.14} & 0.31 & 0.23 & \textbf{0.15} & \textbf{0.11} & 1.24 & 1.04 & 0.87 & 0.74 & 0.31 & 0.36 & 0.41 & 0.46 \\
\karldagger (reproduction, mean $\pm$ std) & 0.33 & 0.26 & \textbf{0.19} & 0.15 & 0.31 & 0.22 & \textbf{0.15} & \textbf{0.11} & 1.24 & 1.04 & 0.87 & 0.74 & 0.31 & 0.36 & 0.41 & 0.46 \\[-2pt]
& {\scriptsize$\pm$.001} & {\scriptsize$\pm$.002} & {\scriptsize$\pm$.001} & {\scriptsize$\pm$.001} & {\scriptsize$\pm$.002} & {\scriptsize$\pm$.002} & {\scriptsize$\pm$.000} & {\scriptsize$\pm$.000} & {\scriptsize$\pm$.003} & {\scriptsize$\pm$.009} & {\scriptsize$\pm$.007} & {\scriptsize$\pm$.004} & {\scriptsize$\pm$.001} & {\scriptsize$\pm$.006} & {\scriptsize$\pm$.002} & {\scriptsize$\pm$.001} \\
\midrule
\rowcolor{gistrow}
\ourmethod    & 0.33 & 0.26 & \textbf{0.19} & \textbf{0.14} & 0.30 & 0.22 & \textbf{0.15} & \textbf{0.11} & 1.23 & 1.03 & 0.88 & 0.75 & 0.31 & 0.36 & 0.41 & 0.46 \\
\rowcolor{gistrow}
\ourmethodone  & 0.33 & 0.26 & \textbf{0.19} & 0.15 & 0.30 & 0.22 & \textbf{0.15} & \textbf{0.11} & 1.24 & 1.04 & 0.87 & 0.75 & 0.31 & 0.36 & 0.41 & 0.46 \\
\bottomrule
\end{tabular}%
}
\end{table}

%% file: sections/appendix/comparison_tokenizers.tex
\clearpage
\section{Comparison to Independently Trained Tokenizers}
\label{app:comparison_tokenizers}
In this section, we situate \ourmethod relative to existing tokenizers, which differ from it on several axes at once: training data, scale, architecture, and objective.
This comparison therefore serves to contextualize \ourmethod rather than rank it against them.
Because most existing tokenizers provide no comparable mechanism for controlling the encoding through a target error, we instead vary the \textit{input token budget} $T$ for each method, as in \cref{app:app_token_budget_sweep_metrics}.

We re-run each method in our own pipeline.
To this end, we evaluate every method on the \textit{same} ImageNet-100 validation set; only the \textit{training data} differs (noted per row).
For ALIT (a recurrent tokenizer that allocates 32 tokens per iteration), we use its reconstructions from rollout iterations $\{1, 2, 4, 8\}$, i.e. budgets $\{32, 64, 128, 256\}$, and disable its early-halting mechanism so that each budget uses its full token count.
For FlexTok, we draw a single fixed-seed sample from its rectified-flow decoder with the default sampler ($25$ steps, default guidance scale); for One-D-Piece, we use its deterministic decoder.

The results are shown in \cref{tab:comparison_tokenizers}.
Within the controlled block, \ourmethod and \ourmethodone perform on par with \karldagger: every difference is at most $0.01$, i.e., within the seed spread of \karldagger's three retrains (\cref{tab:karl_reproduction}).
Furthermore, despite the reproduction gap detailed in \cref{app:karl_reproduction}, \ourmethod remains competitive with ALIT, the strongest independently trained tokenizer that does not share our baseline's method.
For budgets $\geq$~128 tokens, it matches or surpasses ALIT on L1, LPIPS, and SSIM (e.g. at $256$ tokens, LPIPS $0.14$ vs. $0.16$ and SSIM $0.46$ vs. $0.44$).

Lastly, FlexTok and One-D-Piece generally underperform on LPIPS, L1, and SSIM compared to the other tokenizers.
We attribute this gap to the metrics' focus on per-image reconstruction fidelity, which inherently favors reconstruction-trained tokenizers (GIST, KARL, and ALIT).
In contrast, FlexTok and One-D-Piece trade pixel fidelity for generative realism by using a rectified-flow decoder and a perceptually trained decoder, respectively.
Consistent with this trade-off, both methods remain competitive on DreamSim.

We omit rFID --- the metric that would best capture this generative advantage --- for two reasons.
First, rFID is unpaired: it compares the \textit{distribution} of reconstructions to that of the inputs, so a realistic reconstruction of the wrong instance (e.g. a plausible but different bird) is not penalized, whereas the per-image metrics above measure exactly this failure.
Since our targets specify per-image error bounds, per-image fidelity is the quantity of interest; \citet{duggal2025karl} make the same argument for prioritizing per-image metrics over FID.
Second, FID estimates are biased and high-variance at small sample sizes~\citep{chong2020effectively}, and the 5{,}000 images of our ImageNet-100 validation set fall well below the regime where they become reliable.
We note that \cref{tab:comparison_tokenizers} therefore does not reflect the generative realism that FlexTok and One-D-Piece are optimized for, for which we refer the reader to the rFID results in their respective papers~\citep{bachmann2025flextok,miwa2025onedpiece}.

\begin{table}[H]
\centering
\caption{\textbf{Reconstruction fidelity of adaptive tokenizers across token budgets on ImageNet-100.}
Top block: models we train under one identical protocol (our controlled comparison); for \karldagger, we report the mean over its three retrains (\cref{tab:karl_reproduction}).
Bottom block: Independently trained tokenizers, re-evaluated by us under the same protocol and reported for reference only.
They differ from the controlled block in architecture, training objective, training data, and scale.
In the top block, all methods perform comparably because \ourmethod's advantage over the baselines comes only from varying its two targets, which are both fixed at $0$ here.
In the bottom block, the official KARL checkpoint performs best across most metrics.}
\label{tab:comparison_tokenizers}
\small
\setlength{\tabcolsep}{3pt}
\resizebox{\textwidth}{!}{%
\begin{tabular}{l l cccc cccc cccc cccc}
\toprule
& & \multicolumn{4}{c}{LPIPS $\downarrow$}
& \multicolumn{4}{c}{DreamSim $\downarrow$}
& \multicolumn{4}{c}{L1 Loss $\times$10 $\downarrow$}
& \multicolumn{4}{c}{SSIM $\uparrow$} \\
\cmidrule(lr){3-6}\cmidrule(lr){7-10}\cmidrule(lr){11-14}\cmidrule(lr){15-18}
Method & Train & 32 & 64 & 128 & 256 & 32 & 64 & 128 & 256 & 32 & 64 & 128 & 256 & 32 & 64 & 128 & 256 \\
\midrule
\multicolumn{18}{l}{\textit{Controlled: trained by us under identical conditions}} \\
\karldagger   & IN-100 & \textbf{0.33} & \textbf{0.26} & \textbf{0.19} & 0.15 & 0.31 & \textbf{0.22} & \textbf{0.15} & \textbf{0.11} & 1.24 & 1.04 & \textbf{0.87} & \textbf{0.74} & \textbf{0.31} & \textbf{0.36} & \textbf{0.41} & \textbf{0.46} \\
\rowcolor{gistrow}
\ourmethodone & IN-100 & \textbf{0.33} & \textbf{0.26} & \textbf{0.19} & 0.15 & \textbf{0.30} & \textbf{0.22} & \textbf{0.15} & \textbf{0.11} & 1.24 & 1.04 & \textbf{0.87} & 0.75 & \textbf{0.31} & \textbf{0.36} & \textbf{0.41} & \textbf{0.46} \\
\rowcolor{gistrow}
\ourmethod    & IN-100 & \textbf{0.33} & \textbf{0.26} & \textbf{0.19} & \textbf{0.14} & \textbf{0.30} & \textbf{0.22} & \textbf{0.15} & \textbf{0.11} & \textbf{1.23} & \textbf{1.03} & 0.88 & 0.75 & \textbf{0.31} & \textbf{0.36} & \textbf{0.41} & \textbf{0.46} \\
\midrule
\multicolumn{18}{l}{\textit{Reference: independently trained --- not directly comparable}} \\
KARL (public ckpt., our eval)              & IN-100 & \textbf{0.30} & \textbf{0.24} & \textbf{0.19} & \textbf{0.14} & 0.28 & 0.21 & 0.16 & 0.11 & \textbf{1.14} & \textbf{0.98} & \textbf{0.80} & \textbf{0.70} & \textbf{0.34} & \textbf{0.38} & \textbf{0.45} & \textbf{0.49} \\
ALIT~\citep{duggal2024alit}              & IN-100 & 0.31 & 0.25 & 0.20 & 0.16 & 0.26 & 0.20 & 0.14 & 0.10 & 1.18 & 1.01 & 0.88 & 0.76 & 0.32 & 0.36 & 0.40 & 0.44 \\
FlexTok-d12~\citep{bachmann2025flextok}  & IN-1K  & 0.42 & 0.35 & 0.27 & 0.21 & 0.27 & 0.22 & 0.17 & 0.14 & 1.53 & 1.33 & 1.05 & 0.85 & 0.27 & 0.30 & 0.38 & 0.45 \\
FlexTok-d18/d18~\citep{bachmann2025flextok} & IN-1K & 0.37 & 0.31 & 0.25 & 0.19 & \textbf{0.23} & 0.19 & 0.16 & 0.12 & 1.37 & 1.21 & 1.02 & 0.84 & 0.30 & 0.33 & 0.38 & 0.45 \\
FlexTok-d18/d28~\citep{bachmann2025flextok} & IN-1K & 0.38 & 0.35 & 0.27 & 0.20 & 0.24 & 0.22 & 0.17 & 0.12 & 1.39 & 1.31 & 1.05 & 0.84 & 0.29 & 0.31 & 0.37 & 0.45 \\
One-D-Piece-S~\citep{miwa2025onedpiece}  & IN-1K  & 0.36 & 0.28 & 0.21 & 0.19 & 0.29 & 0.20 & 0.14 & 0.12 & 1.35 & 1.17 & 1.02 & 0.94 & 0.32 & 0.35 & 0.38 & 0.40 \\
One-D-Piece-B~\citep{miwa2025onedpiece}  & IN-1K  & 0.35 & 0.27 & 0.21 & 0.18 & 0.27 & 0.18 & 0.13 & 0.11 & 1.35 & 1.17 & 1.01 & 0.92 & 0.32 & 0.35 & 0.38 & 0.41 \\
One-D-Piece-L~\citep{miwa2025onedpiece}  & IN-1K  & 0.35 & 0.26 & 0.20 & 0.17 & 0.25 & \textbf{0.17} & \textbf{0.12} & \textbf{0.09} & 1.33 & 1.14 & 0.96 & 0.88 & 0.33 & 0.35 & 0.39 & 0.41 \\
\bottomrule
\end{tabular}%
}
\end{table}

%% file: sections/appendix/tokens_vs_quality_metrics.tex
\clearpage
\section{Tokens Used vs.\ Reconstruction Quality Metrics}
\label{app:app_tokens_vs_quality_metrics}

\begin{figure}[H]
    \centering
    \includegraphics[width=\textwidth]{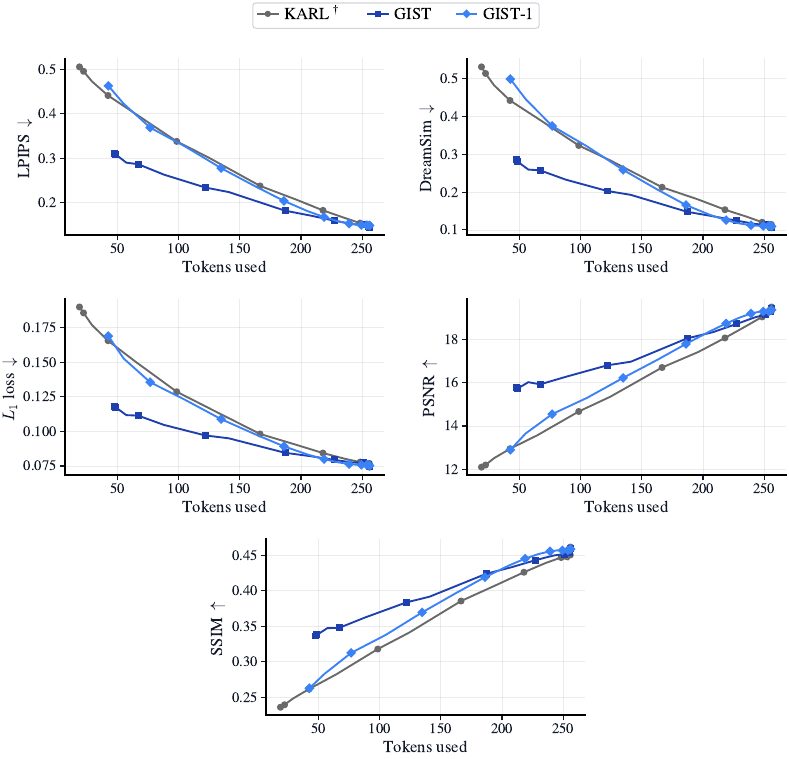}
    \caption{\textbf{Tokens vs.\ reconstruction quality metrics on ImageNet-100.}
    Pareto fronts on the ImageNet-100 validation set for the full set of reconstruction quality metrics ($L_1$, LPIPS, SSIM, PSNR, and DreamSim), obtained by varying $\lrec$ for \karldagger and \ourmethod (fixing $\lsem = 0$), and by varying $\lsem$ for \ourmethodone.
    \ourmethod outperforms \karldagger on every metric at any token count.
    }
    \label{fig:app_tokens_vs_metrics}
\end{figure}

We complement the LPIPS and DreamSim Pareto fronts of \cref{fig:tokens_vs_lpips_dreamsim_pareto} with the full set of reconstruction quality metrics ($L_1$, LPIPS, SSIM, PSNR, and DreamSim) in \cref{fig:app_tokens_vs_metrics}.
Each panel compares \ourmethod, \ourmethodone, and \karldagger along the token-fidelity Pareto front in a different metric on the ImageNet-100 validation set, obtained by varying $\lrec$ over its full range for \karldagger and \ourmethod (fixing $\lsem = 0$), and by varying $\lsem$ for \ourmethodone.
The conclusions drawn from \cref{fig:tokens_vs_lpips_dreamsim_pareto} hold across all metrics.
\ourmethod achieves better reconstruction quality than \karldagger at any number of active tokens, and the gap widens as images are compressed into fewer tokens.

%% file: sections/appendix/coco_evaluation.tex
\clearpage
\section{Evaluation on Multi-Object Scenes (MS COCO)}
\label{app:coco_evaluation}
To test whether the reconstruction gains extend beyond the object-centric images of ImageNet-100, we repeat the comparison of \cref{fig:tokens_vs_lpips_dreamsim_pareto} on MS COCO~\citep{lin2014coco}.
Its everyday scenes contain multiple objects in context, allowing us to examine both generalization to a different image distribution and the effect of scene complexity.

\paragraph{Setup.}
We evaluate on the $5{,}000$ images of the COCO 2017 validation set, which contain an average of $7.4$ annotated object instances per image, with $87\%$ containing at least two.
All models use their ImageNet-100-trained checkpoints without further training or adaptation.
To retain more scene context, we resize each image to $256$ pixels on its shorter side and subsequently apply a $256 \times 256$ center crop.
As in \cref{fig:tokens_vs_lpips_dreamsim_pareto}, we fix the input token budget at $T = 256$ and vary $\lrec$ for \karldagger and \ourmethod (fixing $\lsem = 0$), and $\lsem$ for \ourmethodone.
We report LPIPS and DreamSim at matched mean active token counts.

\begin{figure}[H]
    \centering
    \includegraphics[width=\textwidth]{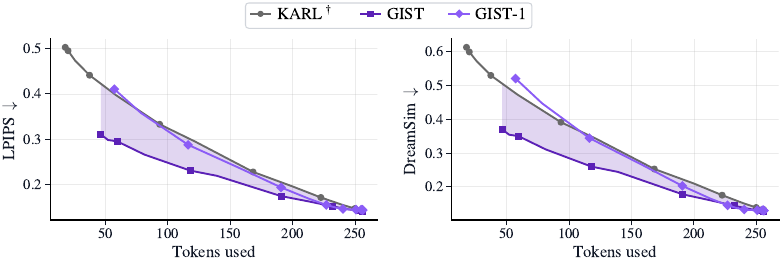}
    \caption{\textbf{The reconstruction gains extend to multi-object scenes.}
    Token-fidelity Pareto fronts on the COCO 2017 validation set, using the ImageNet-100-trained checkpoints without adaptation.
    As in \cref{fig:tokens_vs_lpips_dreamsim_pareto}, we vary $\lrec$ for \karldagger and \ourmethod (fixing $\lsem = 0$), and vary $\lsem$ for \ourmethodone.
    The shaded regions highlight \ourmethod's reduction in reconstruction error relative to \karldagger at matched token counts.
    }
    \label{fig:app_coco_pareto}
\end{figure}

\paragraph{Reconstruction quality.}
\Cref{fig:app_coco_pareto} shows that the findings on ImageNet-100 carry over to COCO.
\ourmethod achieves lower LPIPS and DreamSim than \karldagger at every compared token count, with the largest relative gains under heavy compression: up to $27\%$ lower LPIPS and $28\%$ lower DreamSim.
\ourmethodone again performs on par with \karldagger, suggesting that the reconstruction benefit of jointly controlling pixel-level and semantic fidelity extends to multi-object scenes outside the training distribution.

\begin{figure}[H]
    \centering
    \includegraphics[width=\textwidth]{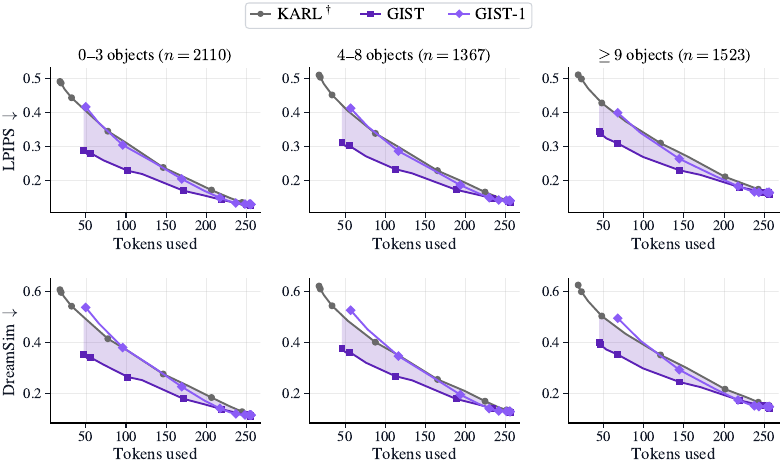}
    \caption{\textbf{The reconstruction advantage persists in crowded scenes.}
    Same evaluation as \cref{fig:app_coco_pareto}, with images grouped by the number of annotated object instances in the original image.
    Each column corresponds to an object-count range, and $n$ denotes the number of images in that group.
    \ourmethod outperforms \karldagger in every group, with the largest relative gains in sparse scenes at low token counts.
    }
    \label{fig:app_coco_density_split}
\end{figure}

\paragraph{Scene complexity.}
To examine how the gains vary with scene complexity, we group the validation images by their number of annotated object instances: $0$--$3$, $4$--$8$, and $\geq 9$.
These counts refer to the original images before cropping and serve as a proxy for scene complexity.
We recompute the Pareto fronts within each group in \cref{fig:app_coco_density_split}.
At the same reconstruction target, both \karldagger and \ourmethod allocate more tokens to scenes containing more objects.
For example, at $\lrec = 0.1$, \karldagger uses an average of $102$ tokens in the sparsest group and $154$ in the densest.

\ourmethod achieves lower LPIPS and DreamSim than \karldagger at matched token counts in all three groups.
The largest relative gains occur in the sparsest scenes: the maximum LPIPS reductions are $31\%$, $27\%$, and $22\%$ from the sparsest to the densest group, with corresponding DreamSim reductions of $30\%$, $27\%$, and $24\%$.
This difference is concentrated in the low-token regime.
At $150$ tokens, the LPIPS reduction is similar across groups ($18$--$19\%$).
Thus, \ourmethod retains its advantage in crowded scenes, although under heavy compression the relative gains are smaller in scenes with more objects than in those with fewer.

%% file: sections/appendix/token_budget_sweep_metrics.tex
\clearpage
\section{Maximum Token Budget vs.\ Reconstruction Quality Metrics}
\label{app:app_token_budget_sweep_metrics}

The comparisons in \cref{fig:tokens_vs_lpips_dreamsim_pareto} and \cref{app:app_tokens_vs_quality_metrics} vary the target errors and let each model decide how many tokens to halt and which ones.
In this section, we complement these comparisons by holding the targets fixed and varying the input token budget $T$ instead.
This isolates the capacity of the underlying tokenizer from the effects of varying its targets.

\begin{figure}[H]
    \centering
    \includegraphics[width=\textwidth]{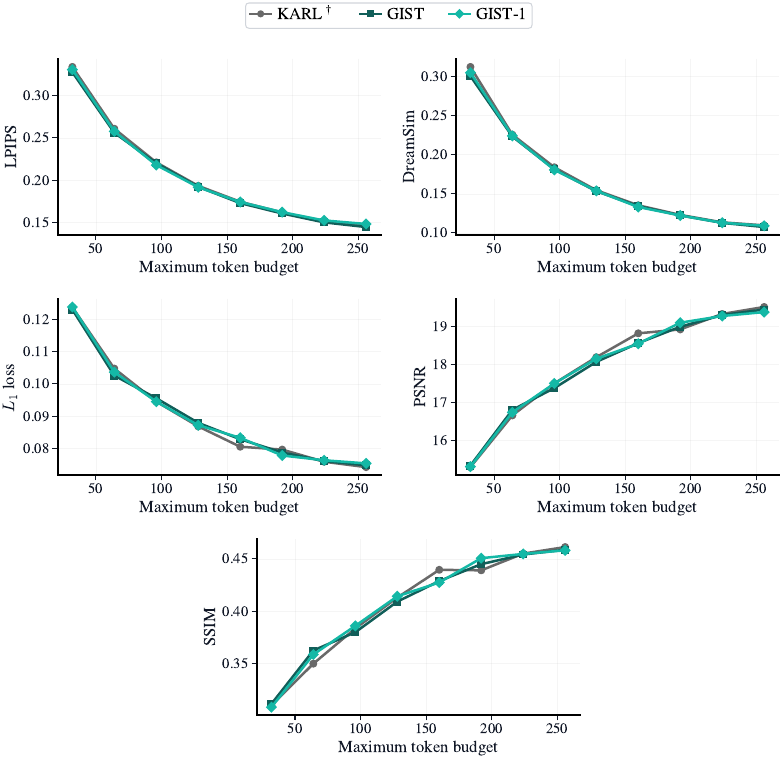}

    \caption{
        \textbf{Reconstruction quality metrics vs.\ input token budget on ImageNet-100.}
        We compare \ourmethod, \ourmethodone, and \karldagger across all five reconstruction quality metrics ($L_1$, LPIPS, SSIM, PSNR, DreamSim) on the ImageNet-100 validation set as we vary the input token budget $T$, with all targets fixed at zero.
        The three models perform on par on every metric, in contrast to the comparison under varying targets (\cref{fig:tokens_vs_lpips_dreamsim_pareto,fig:app_tokens_vs_metrics}), where \ourmethod Pareto-improves reconstruction quality over \karldagger.
    }

    \label{fig:app_token_budget_sweep_metrics}
\end{figure}

In \cref{fig:app_token_budget_sweep_metrics}, we compare \ourmethod, \ourmethodone, and \karldagger on the ImageNet-100 validation set across all five reconstruction quality metrics ($L_1$, LPIPS, SSIM, PSNR, DreamSim) as we vary the input token budget $T$ with all targets held fixed at zero.
\ourmethod, \ourmethodone, and \karldagger perform on par on every metric and at every token budget.
This stands in contrast to the comparison under varying targets, where \ourmethod Pareto-improves reconstruction quality over \karldagger on every metric.

We interpret these results as follows.
All three models share the same architecture, the same VQGAN base, and the same training recipe; they differ only in the target errors that condition the encoder.
Varying $T$ with the targets held fixed thus probes the capacity of the base tokenizer, which semantic conditioning leaves unchanged.
What semantic conditioning changes is how the tokenizer models token complexity.
\karldagger models complexity along a single reconstructive axis: it uses a pixel-wise error target to decide how many tokens to halt and which ones.
\ourmethodone replaces this axis with a semantic one, and \ourmethod models complexity along both.
These axes determine how a model halts tokens as its targets are relaxed.
Under a purely reconstructive axis, tokens are halted by pixel-level redundancy alone, which is blind to the spatial structure that carries an image's meaning.
Under both axes, the semantic target can instead act as a guardrail that protects the image's semantic identity as $\lrec$ is relaxed (\cref{fig:compare_karl_skarl_qualitative}).
The gains of \ourmethod under varying targets therefore stem from how conditioning on these targets organizes the latent space along complementary axes of complexity, rather than from base tokenizer capacity.

%% file: sections/appendix/qualitative_samples.tex
\clearpage
\section{Additional Qualitative \karldagger vs.\ \ourmethod Comparisons}
\label{app:app_qualitative_samples}
We complement the two qualitative examples in the main paper with additional reconstructions from \karldagger and \ourmethod (fixing $\lsem = 0$) at $T = 256$ tokens, organized by image complexity.

\subsection{Synthetic and Abstract Stimuli}
\label{sec:app_qualitative_synthetic}
Synthetic inputs isolate frequency content and color structure without natural-image semantics.

\begin{figure}[H]
    \centering
    \includegraphics[width=0.91\textwidth]{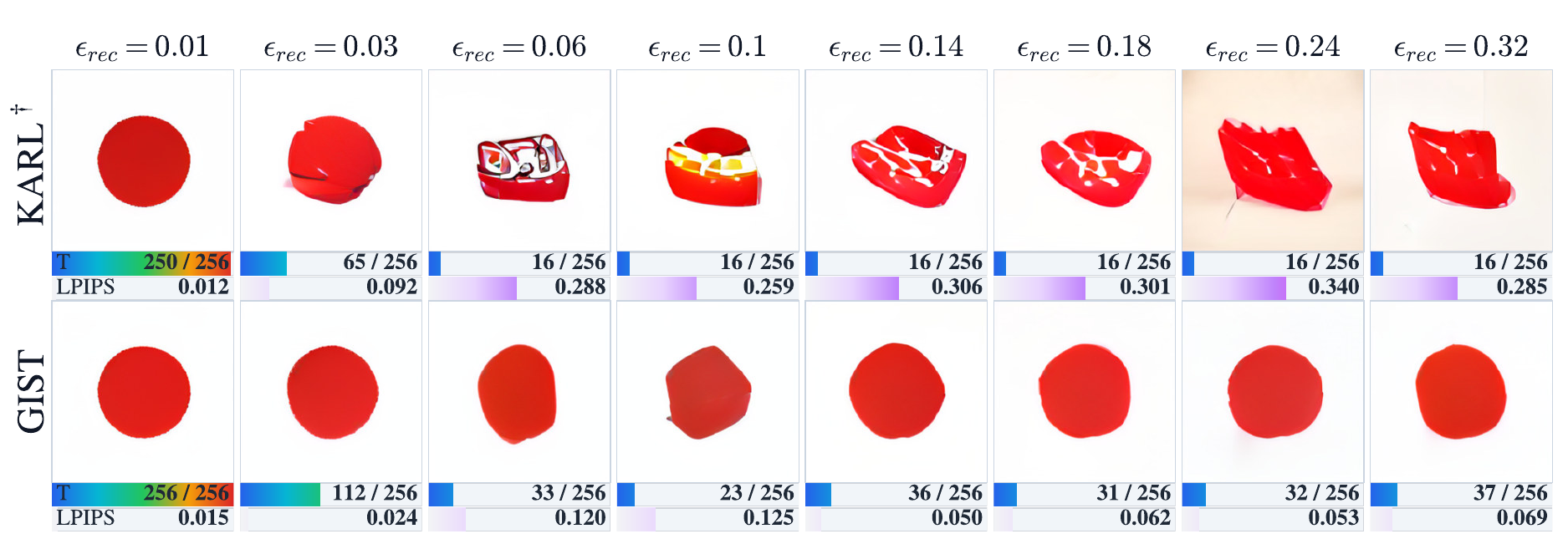}\\[-4pt]
    \includegraphics[width=0.91\textwidth]{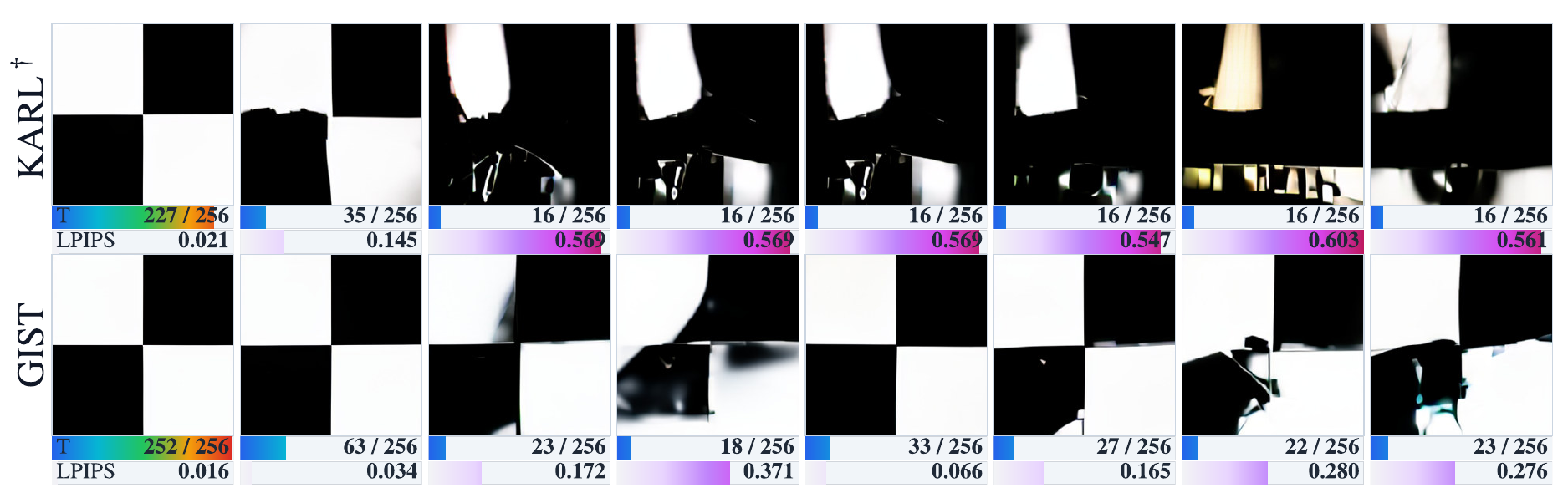}\\[-4pt]
    \includegraphics[width=0.91\textwidth]{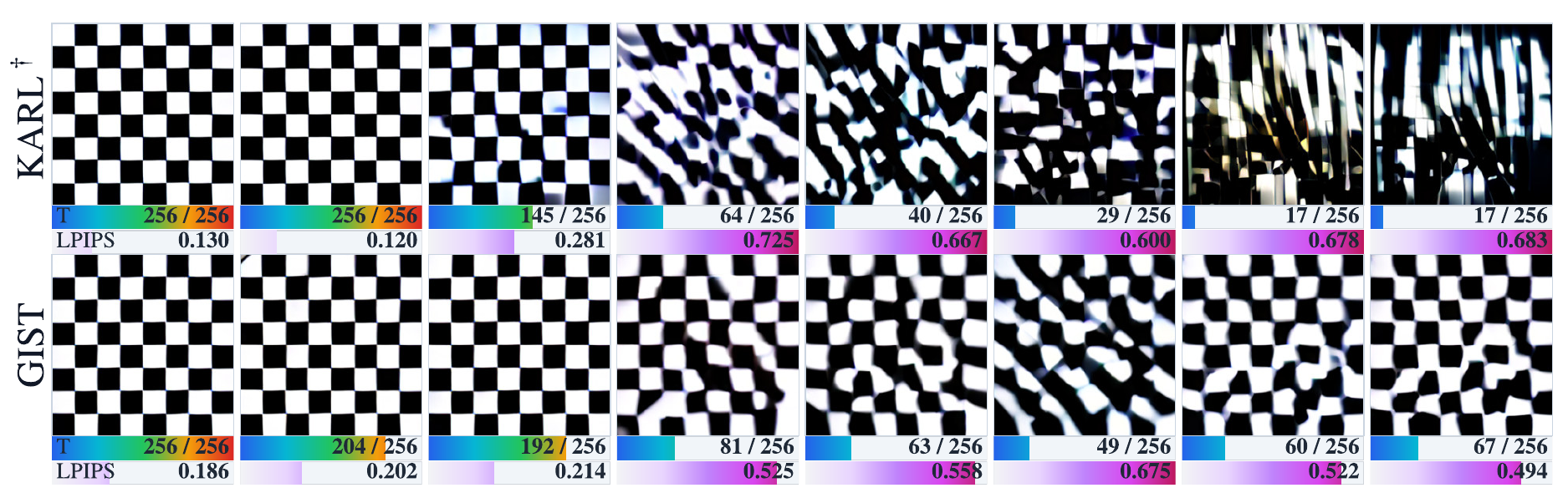}\\[-4pt]
    \includegraphics[width=0.91\textwidth]{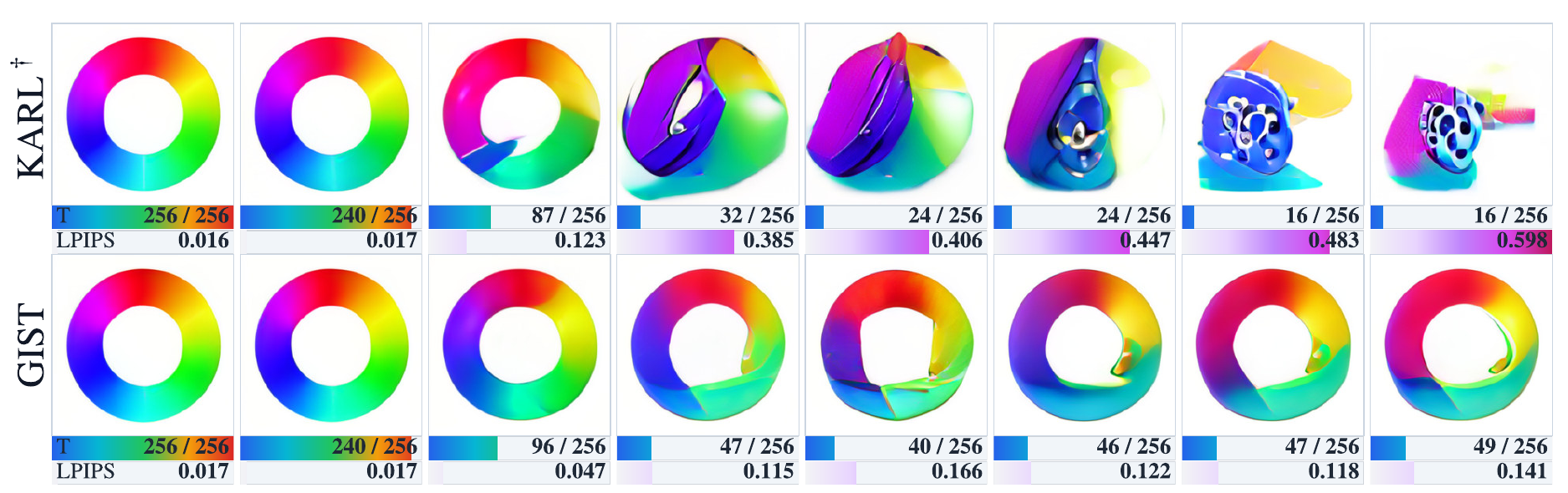}
    \caption{\textbf{Synthetic and abstract stimuli.}
    \karldagger (top of each pair) vs.\ \ourmethod (bottom of each pair) reconstructions across the \lrec spectrum. From top to bottom: \emph{red circle on white}, \emph{checkerboard $2{\times}2$}, \emph{checkerboard $8{\times}8$}, and \emph{color wheel}.}
    \label{fig:app_qual_synthetic}
\end{figure}

\clearpage
\subsection{In-Distribution Samples (ImageNet-100)}
\label{sec:app_qualitative_iid}

We use natural images drawn from classes that lie inside the training distribution.
These probe the typical operating regime of both tokenizers.

\begin{figure}[H]
    \centering
    \includegraphics[width=0.98\textwidth]{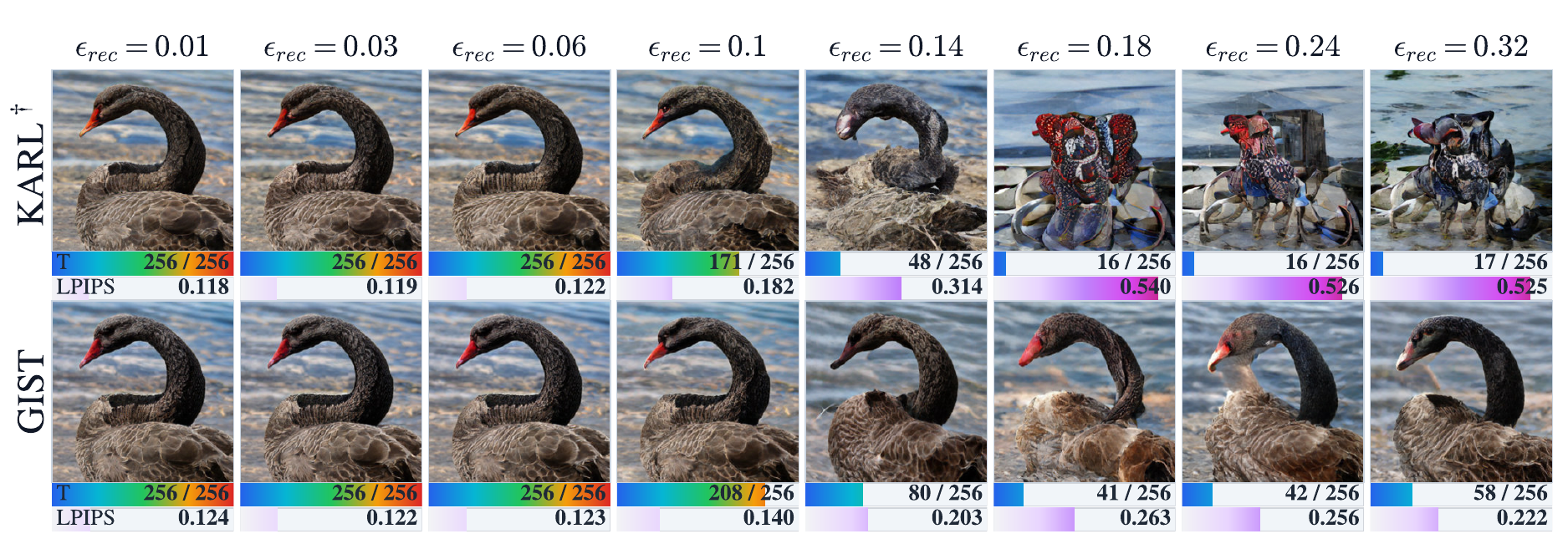}\\[-4pt]
    \includegraphics[width=0.98\textwidth]{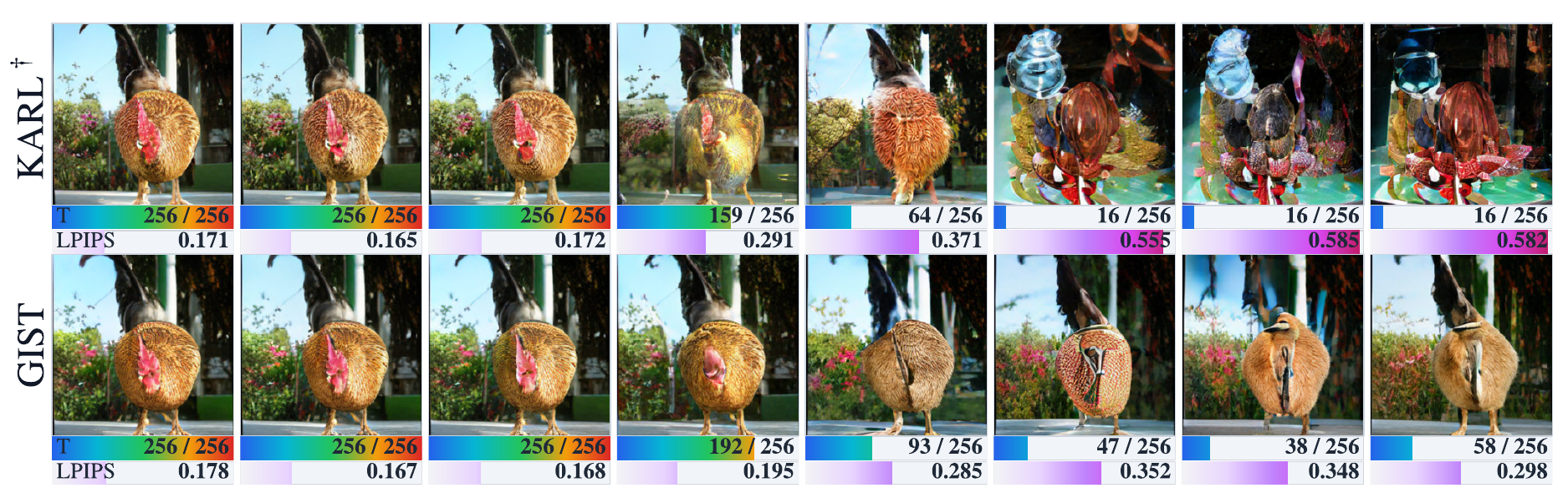}\\[-4pt]
    \includegraphics[width=0.98\textwidth]{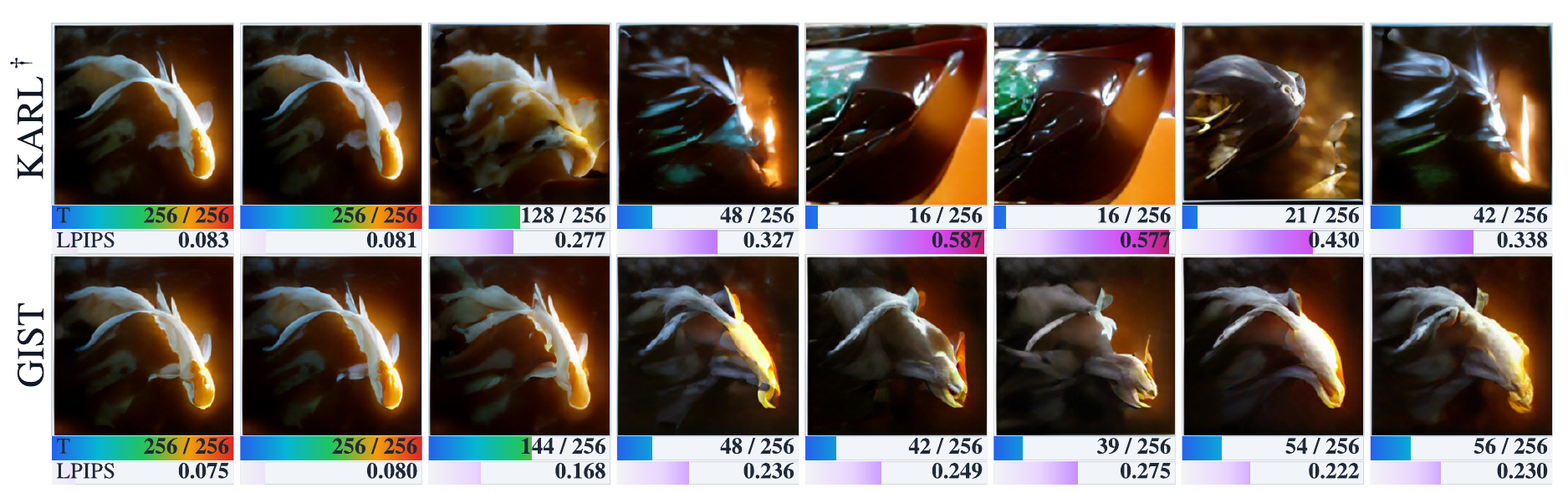}\\[-4pt]
    \includegraphics[width=0.98\textwidth]{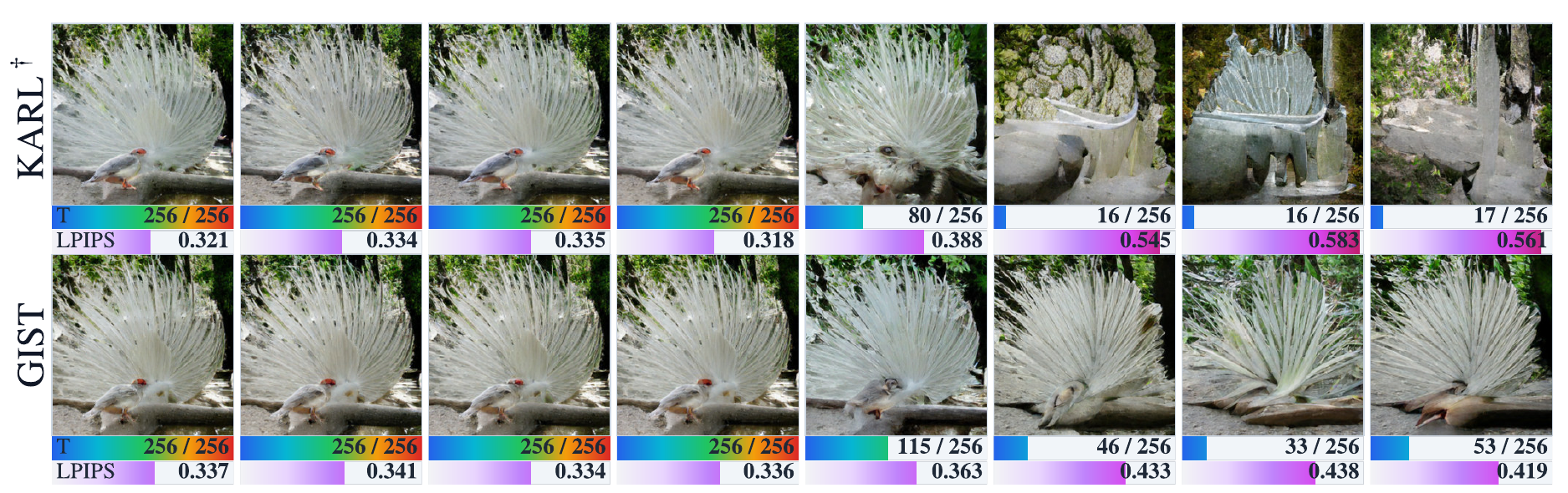}
    \caption{\textbf{In-distribution samples (ImageNet-100).} Same setup as Figure~\ref{fig:app_qual_synthetic}. From top to bottom: \emph{black swan}, \emph{rooster}, \emph{goldfish}, and \emph{peacock}.}
    \label{fig:app_qual_iid}
\end{figure}

\clearpage
\subsection{Out-of-Distribution Natural Photographs}
\label{sec:app_qualitative_ood}

We use real-world photographs of subjects whose classes are not part of the ImageNet-100 training set.
They remain natural-image content, but their specific categories and compositional context lie outside the training distribution.

\begin{figure}[H]
    \centering
    \includegraphics[width=0.98\textwidth]{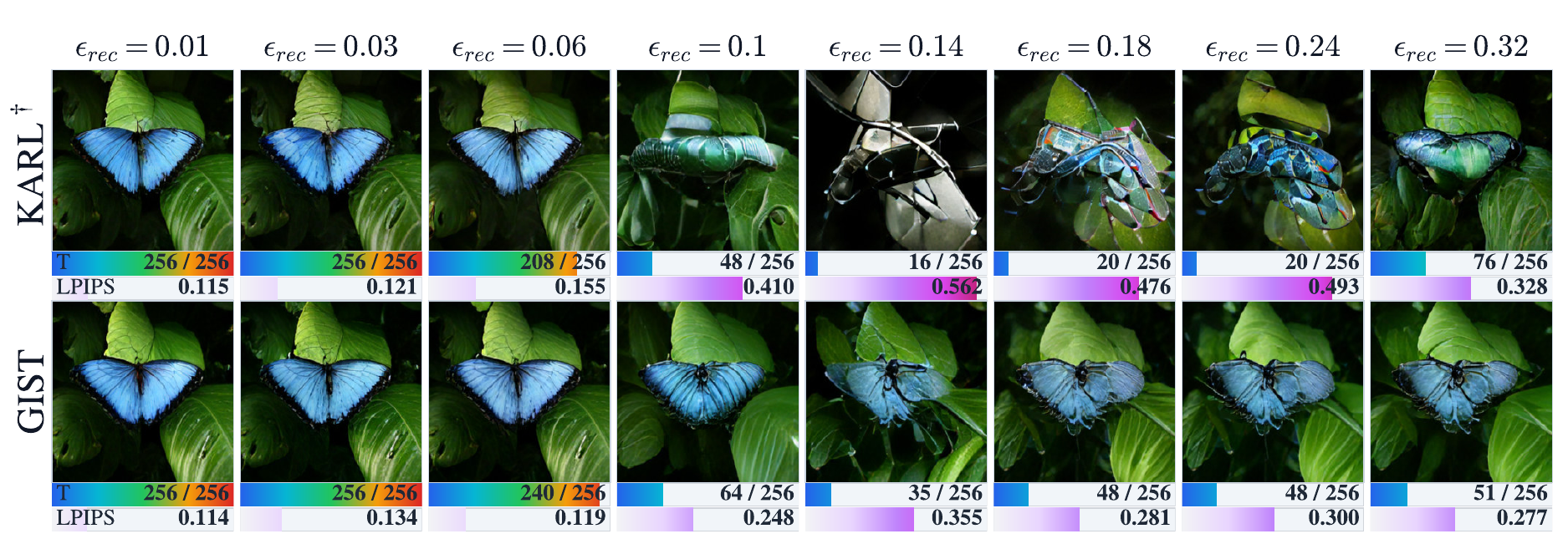}\\[-4pt]
    \includegraphics[width=0.98\textwidth]{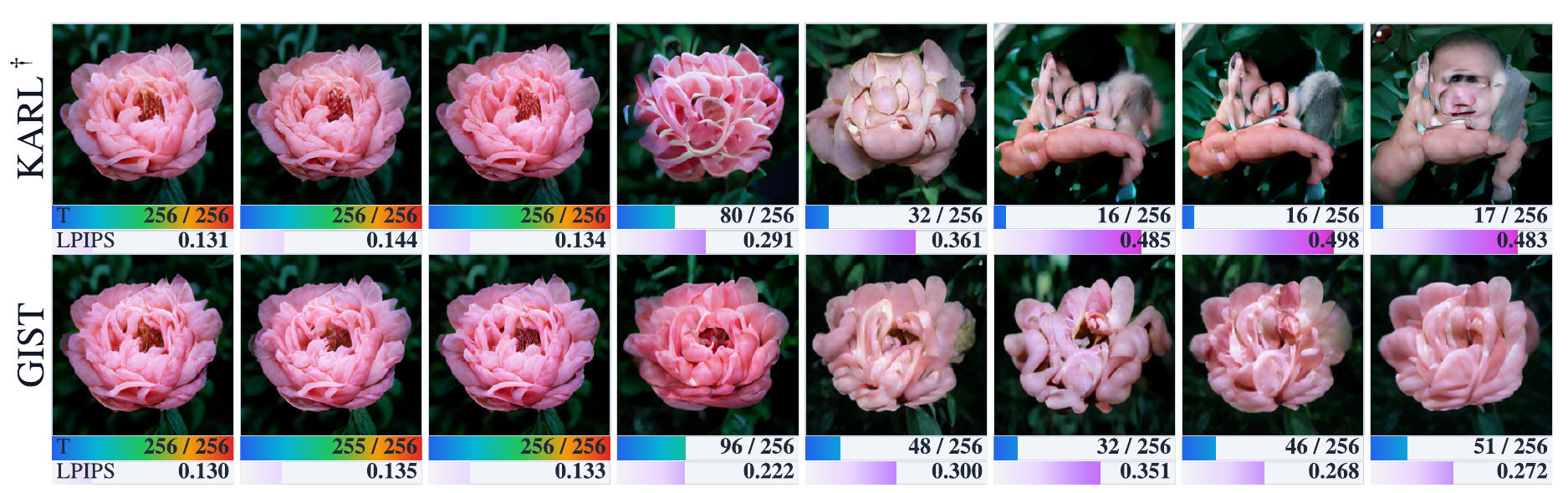}\\[-4pt]
    \includegraphics[width=0.98\textwidth]{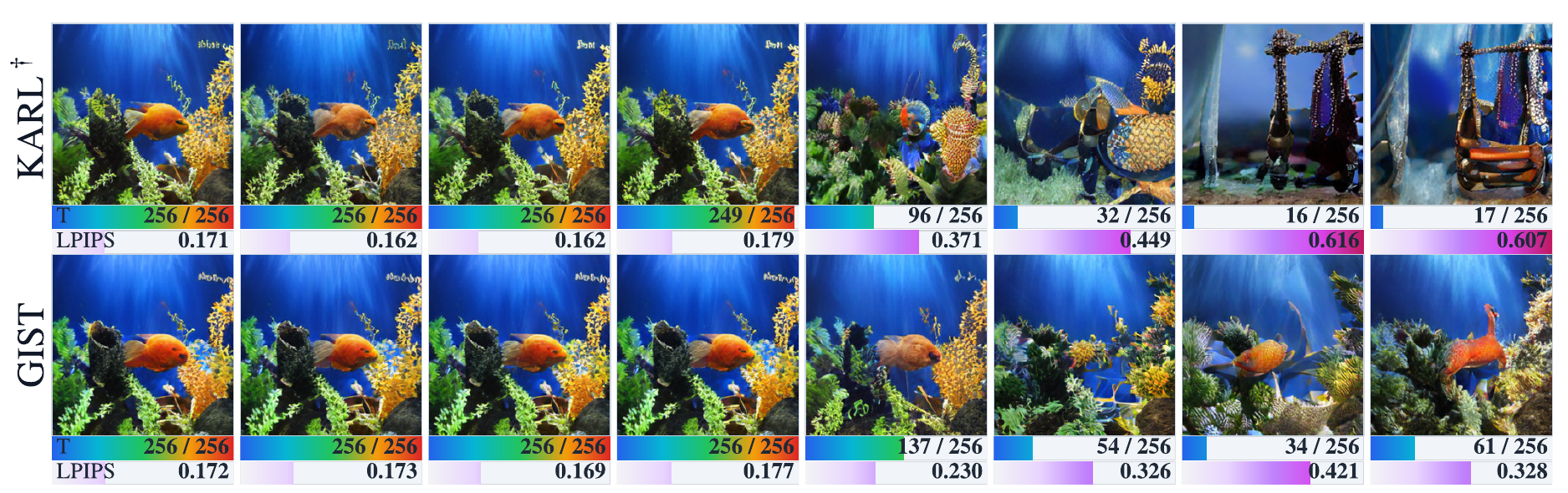}\\[-4pt]
    \includegraphics[width=0.98\textwidth]{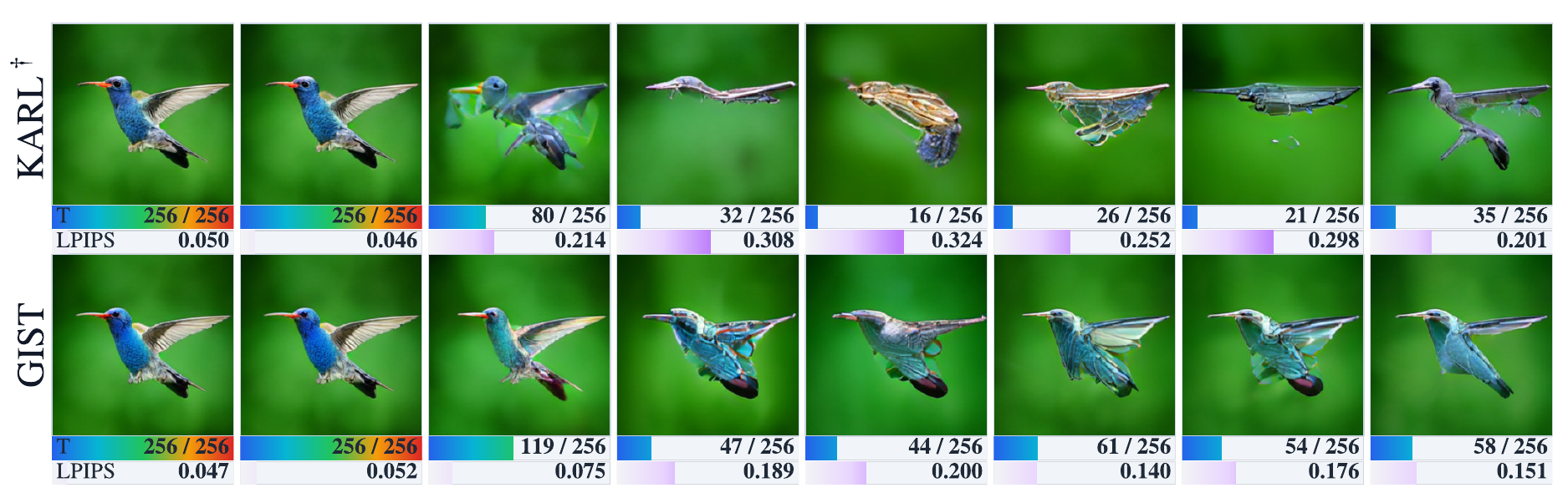}
    \caption{\textbf{Out-of-distribution natural photographs.} Same setup as Figure~\ref{fig:app_qual_synthetic}. From top to bottom: \emph{blue morpho butterfly}, \emph{pink peony}, \emph{goldfish reef}, and \emph{hummingbird}.}
    \label{fig:app_qual_ood}
\end{figure}

\clearpage
\subsection{Difficult Cases: Textures, Fine Detail, and Noise}

\label{sec:app_qualitative_difficult}

Textures, fractals, dense text, and low-frequency noise are dominated by fine-grained detail spread across the entire image rather than by a distinct subject.
Such content is costly to compress, and it probes how both tokenizers behave when no single subject carries the image's identity.

\begin{figure}[H]
    \centering
    \includegraphics[width=0.98\textwidth]{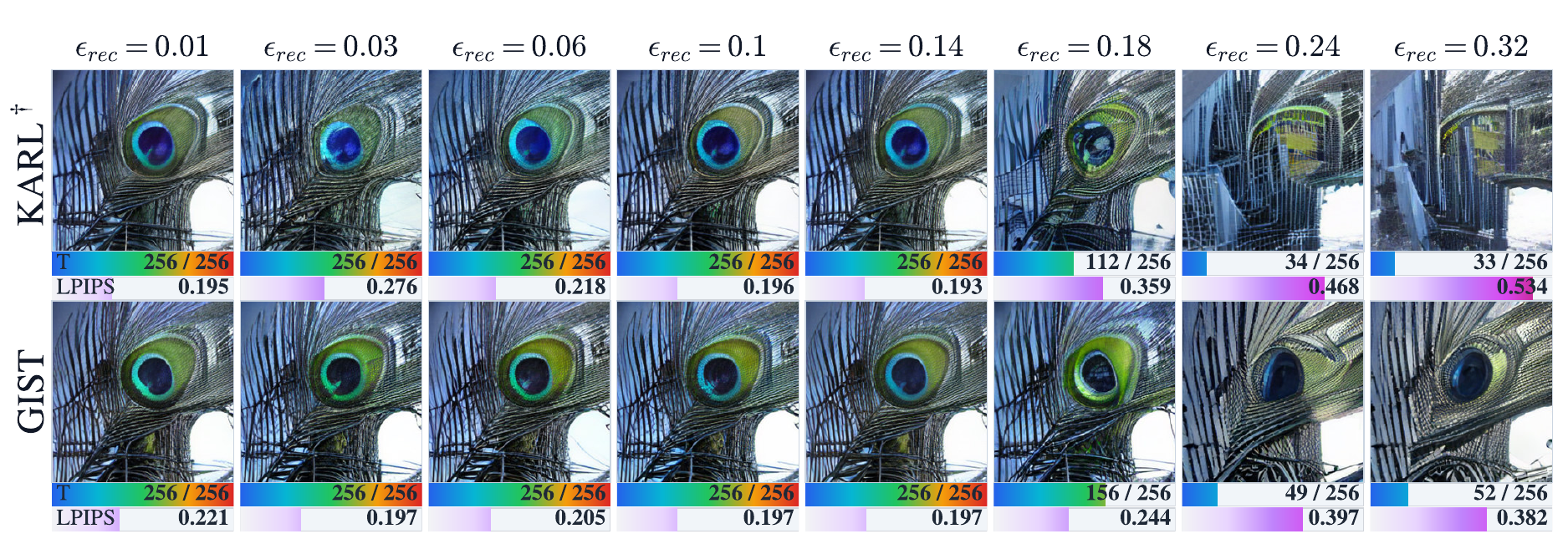}\\[-4pt]
    \includegraphics[width=0.98\textwidth]{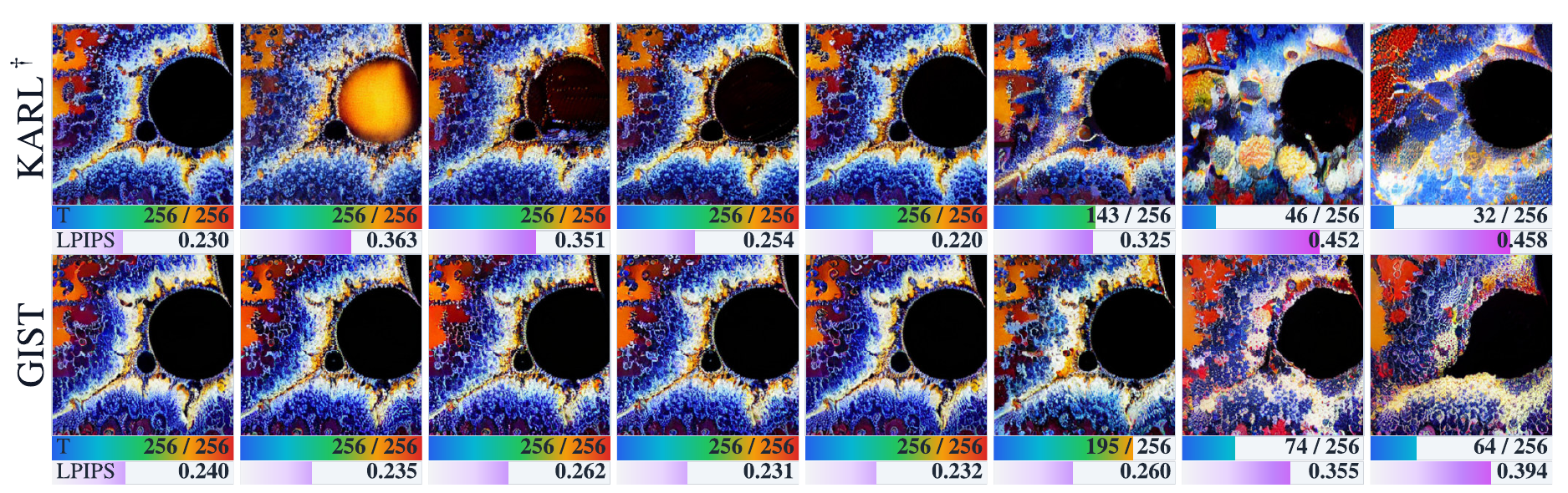}\\[-4pt]
    \includegraphics[width=0.98\textwidth]{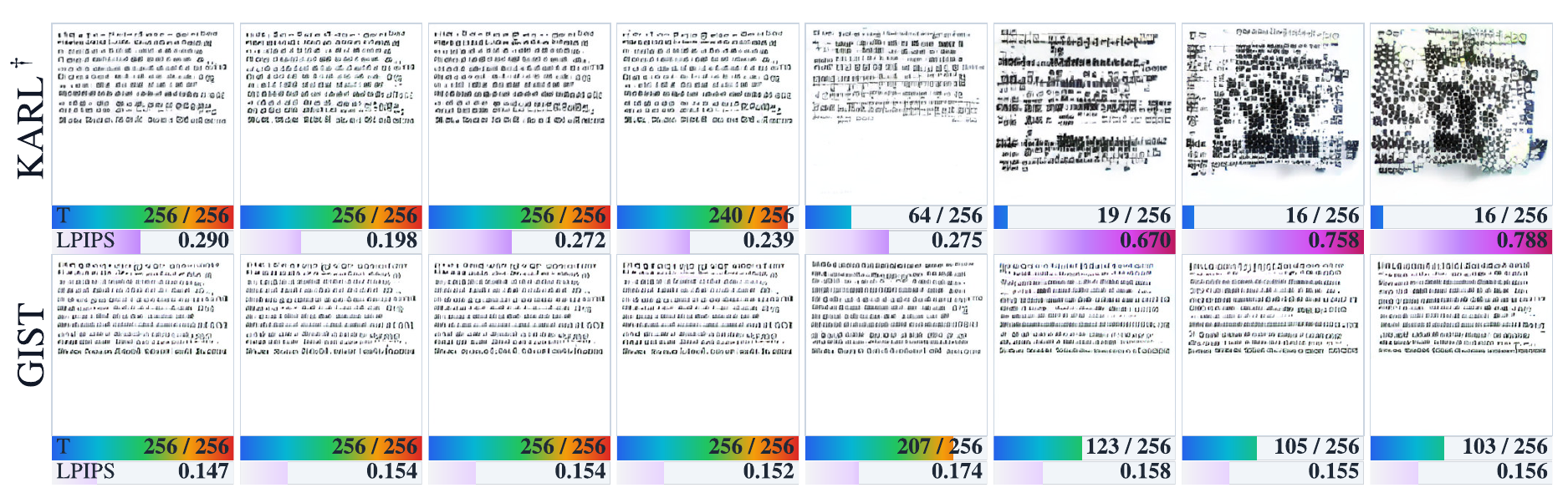}\\[-4pt]
    \includegraphics[width=0.98\textwidth]{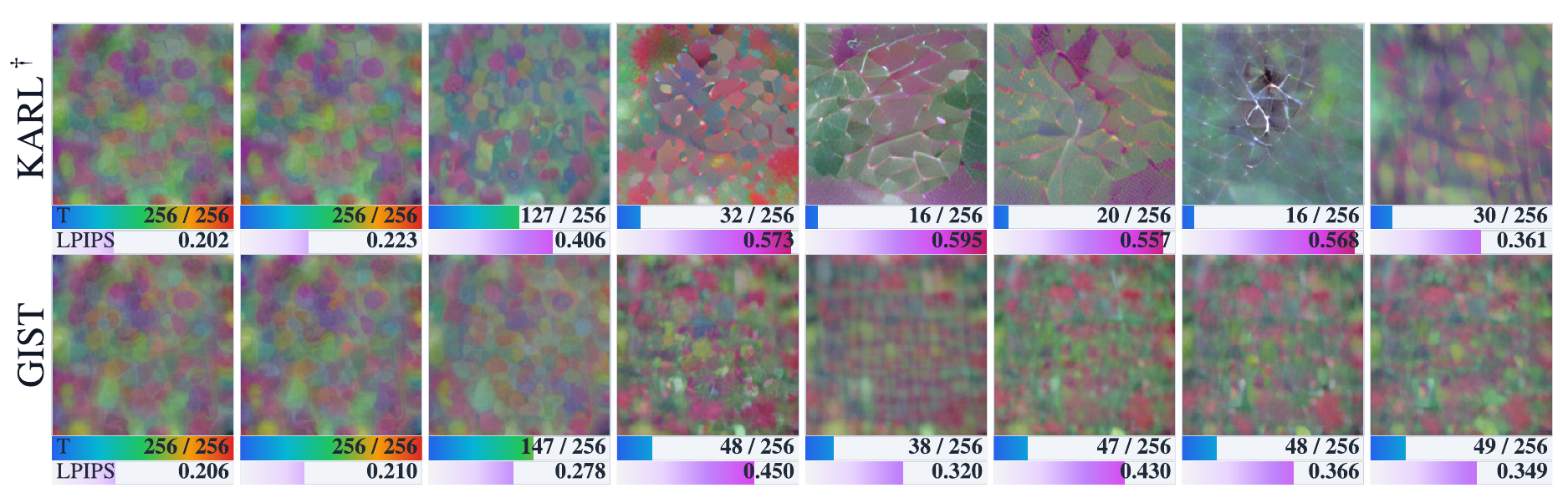}
    \caption{\textbf{Difficult cases: textures and high-frequency detail.} Same setup as Figure~\ref{fig:app_qual_synthetic}. From top to bottom: \emph{peacock feather}, \emph{Mandelbrot fractal}, \emph{text paragraph}, and \emph{low-frequency noise}.}
    \label{fig:app_qual_difficult}
\end{figure}

%% file: sections/appendix/sem_knob_qualitative.tex
\clearpage
\section{Semantic Control: Qualitative Behavior}
\label{app:app_sem_knob_qualitative}

\begin{figure}[H]
    \centering
    \includegraphics[width=0.87\textwidth]{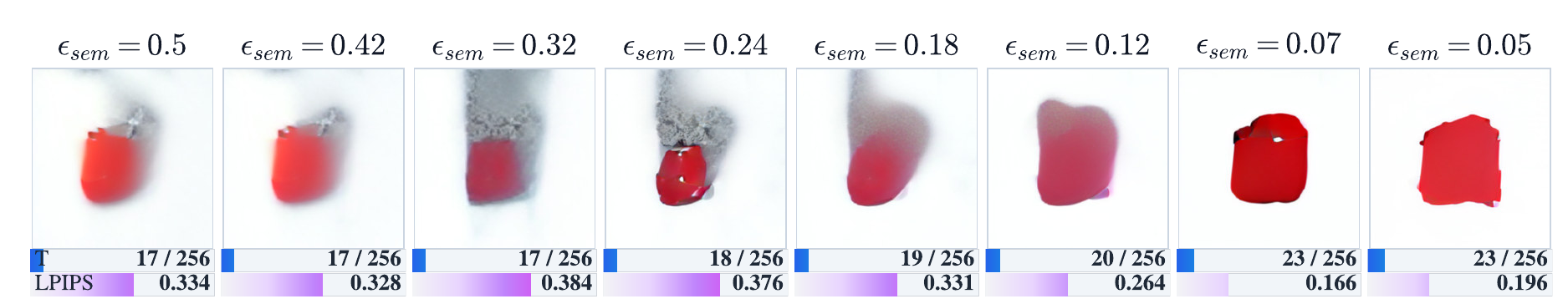}\\[-4pt]
    \includegraphics[width=0.87\textwidth]{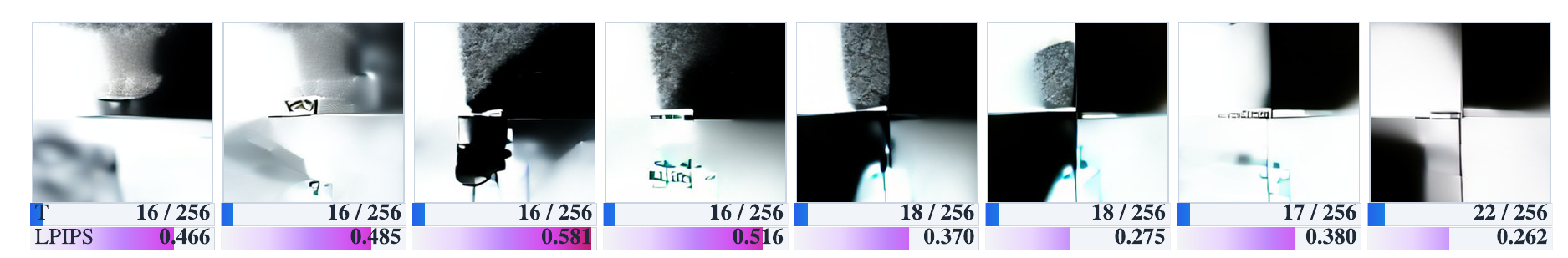}\\[-4pt]
    \includegraphics[width=0.87\textwidth]{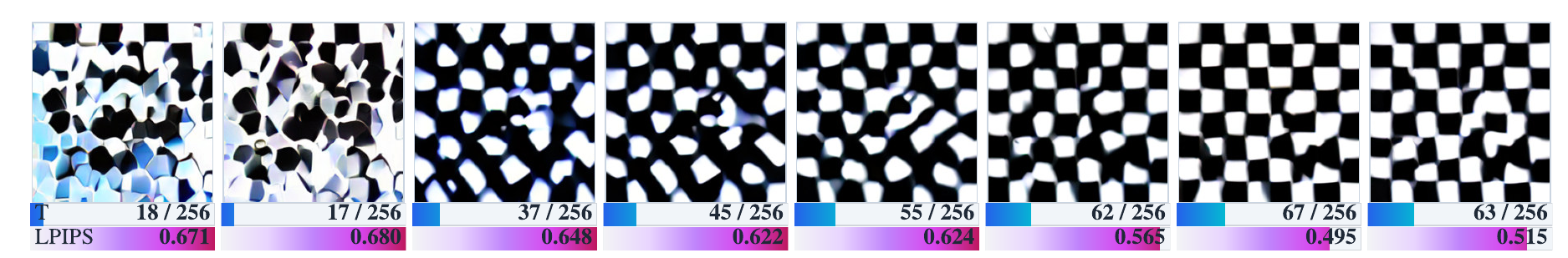}\\[-4pt]
    \includegraphics[width=0.87\textwidth]{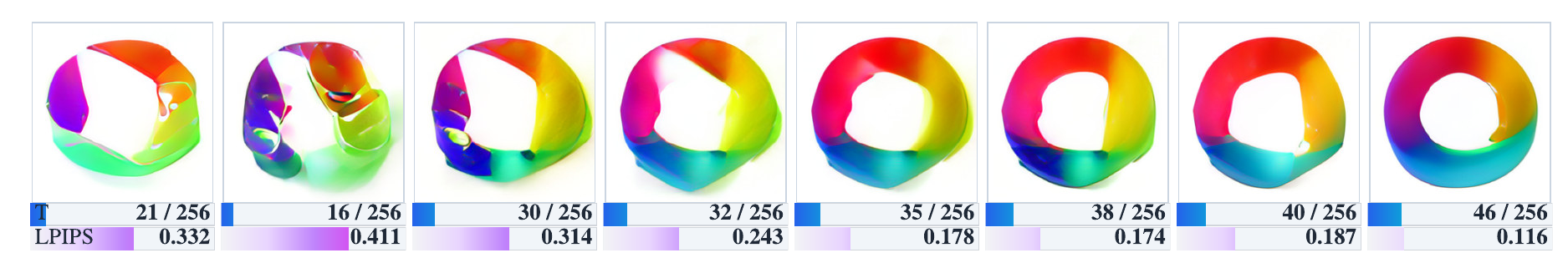}
    \caption{\textbf{Synthetic and abstract stimuli.} \ourmethod reconstructions while varying $\lsem$, with $\lrec = 0.24$ and $T = 256$ tokens. From top to bottom: \emph{red circle on white}, \emph{checkerboard $2{\times}2$}, \emph{checkerboard $8{\times}8$}, and \emph{color wheel}.}
    \label{fig:app_sem_knob_synthetic}
\end{figure}

\begin{figure}[H]
    \centering
    \includegraphics[width=0.87\textwidth]{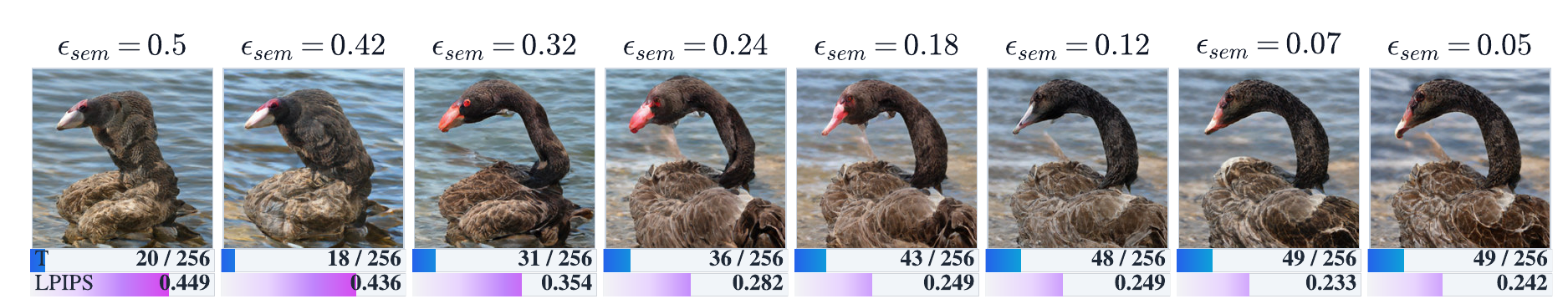}\\[-4pt]
    \includegraphics[width=0.87\textwidth]{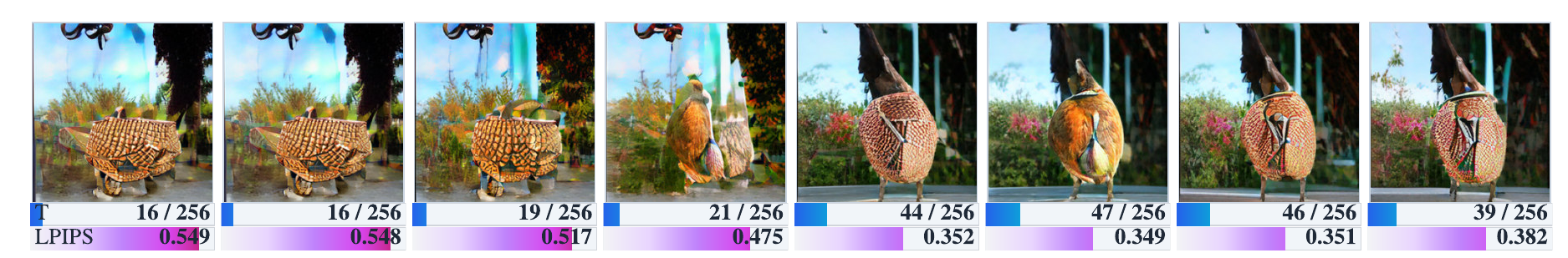}\\[-4pt]
    \includegraphics[width=0.87\textwidth]{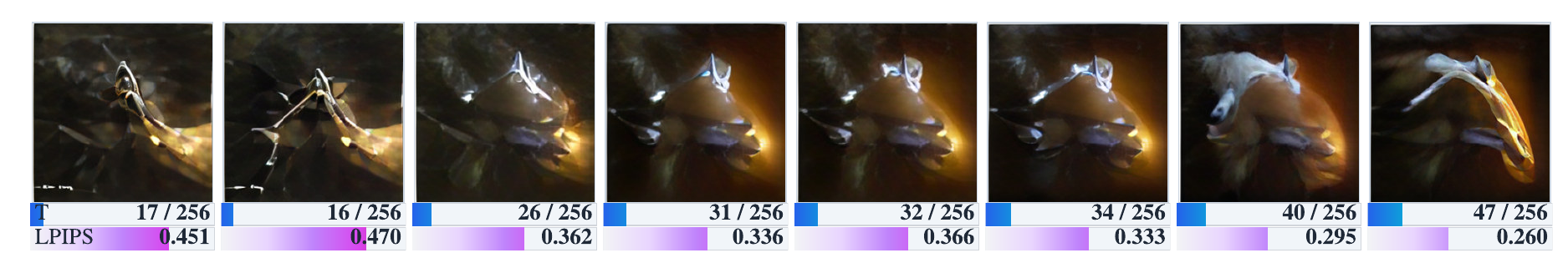}\\[-4pt]
    \includegraphics[width=0.87\textwidth]{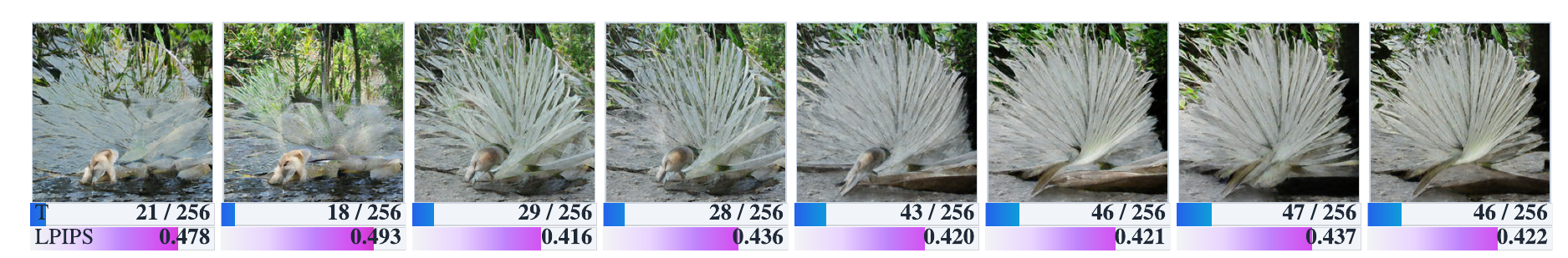}
    \caption{\textbf{In-distribution samples (ImageNet-100).} Same setup as Figure~\ref{fig:app_sem_knob_synthetic}. From top to bottom: \emph{black swan}, \emph{rooster}, \emph{goldfish}, and \emph{peacock}.}
    \label{fig:app_sem_knob_iid}
\end{figure}

\begin{figure}[H]
    \centering
    \includegraphics[width=0.87\textwidth]{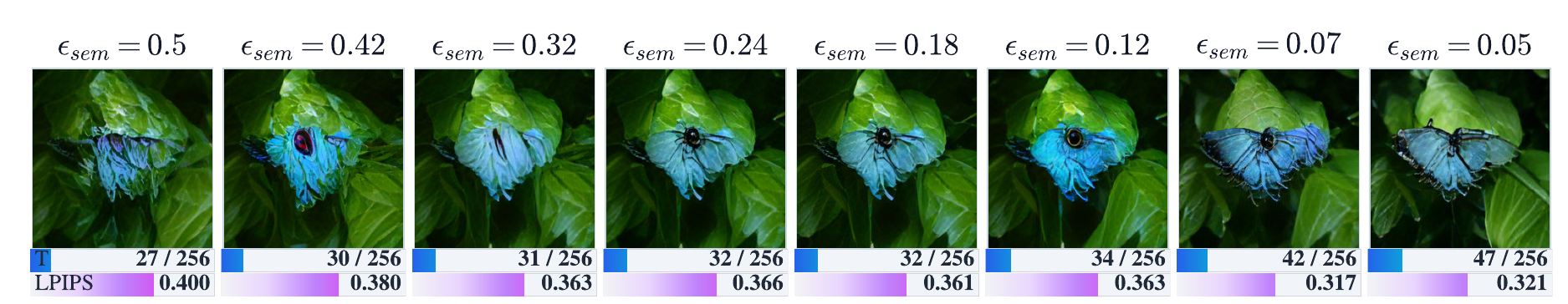}\\[-4pt]
    \includegraphics[width=0.87\textwidth]{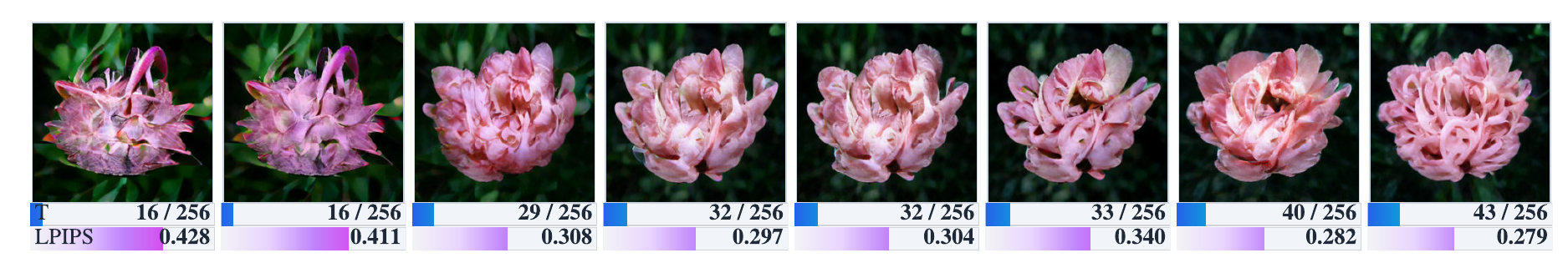}\\[-4pt]
    \includegraphics[width=0.87\textwidth]{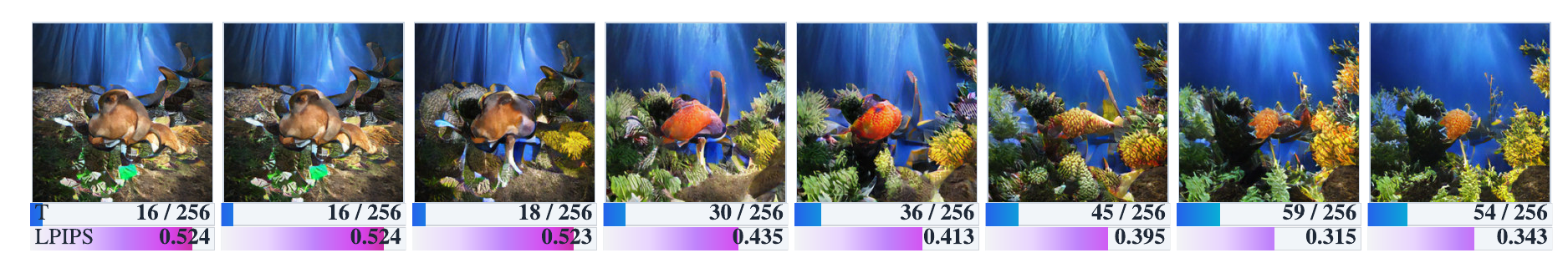}\\[-4pt]
    \includegraphics[width=0.87\textwidth]{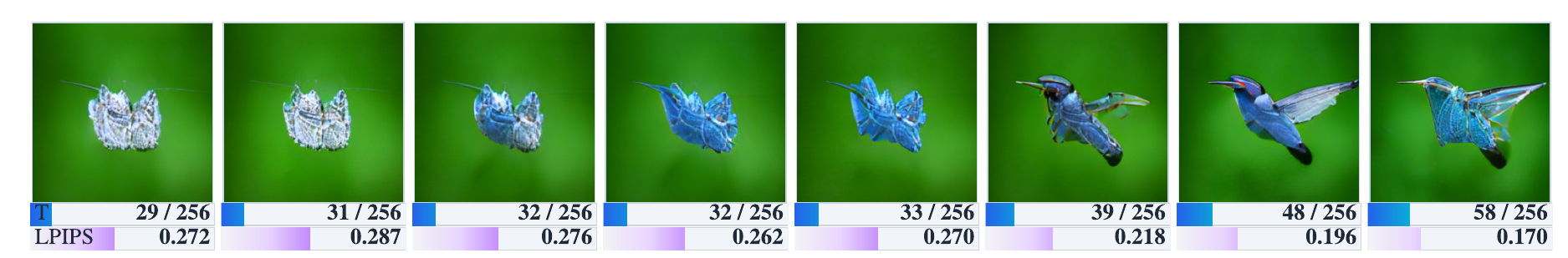}
    \caption{\textbf{Out-of-distribution natural photographs.} Same setup as Figure~\ref{fig:app_sem_knob_synthetic}. From top to bottom: \emph{blue morpho butterfly}, \emph{pink peony}, \emph{goldfish reef}, and \emph{hummingbird}.}
    \label{fig:app_sem_knob_ood}
\end{figure}

\begin{figure}[H]
    \centering
    \includegraphics[width=0.87\textwidth]{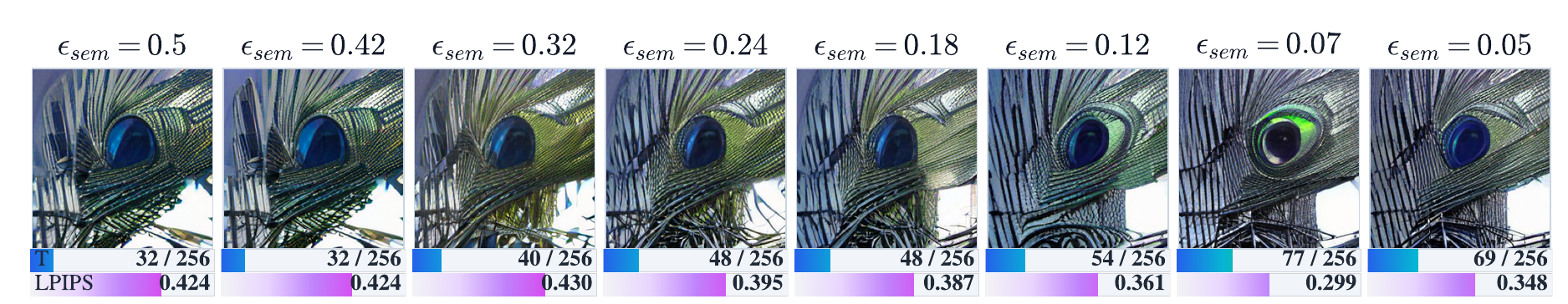}\\[-4pt]
    \includegraphics[width=0.87\textwidth]{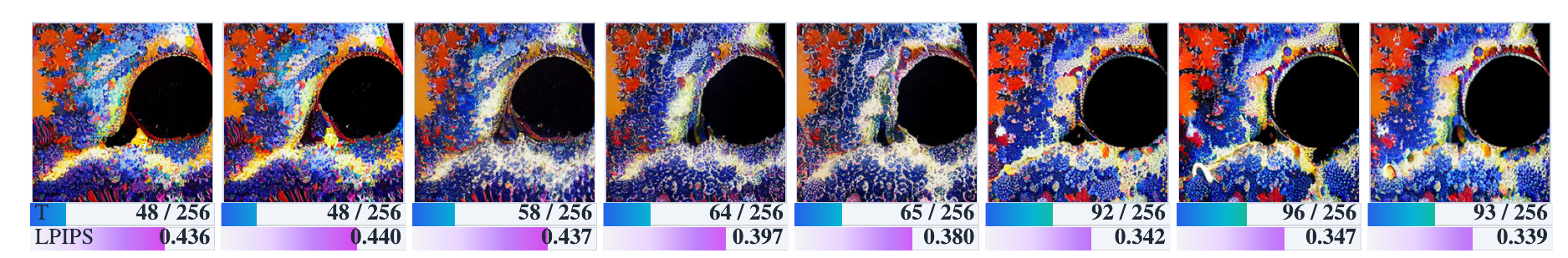}\\[-4pt]
    \includegraphics[width=0.87\textwidth]{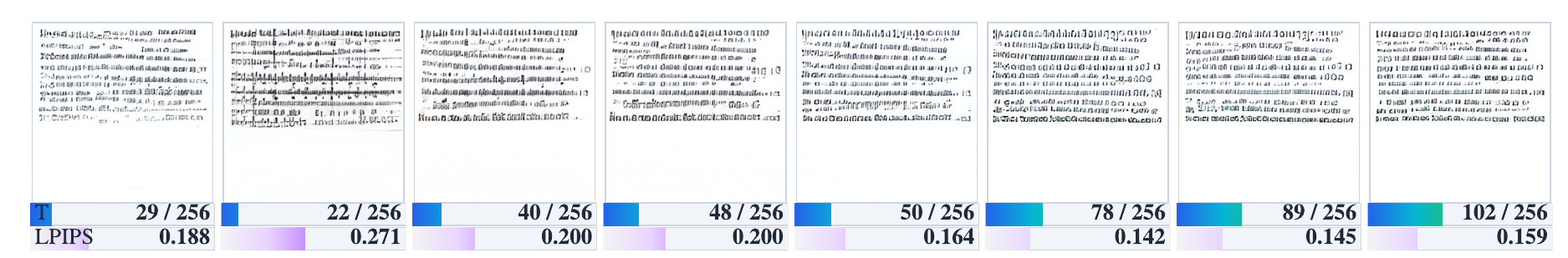}\\[-4pt]
    \includegraphics[width=0.87\textwidth]{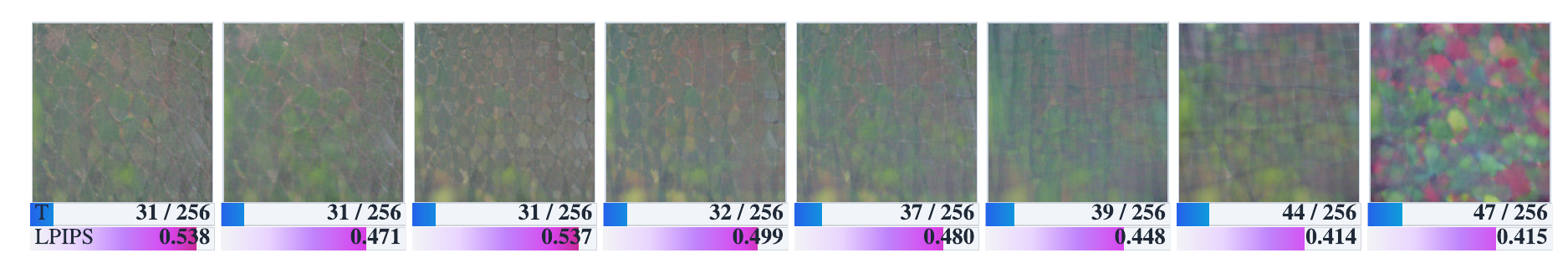}
    \caption{\textbf{Difficult cases: textures and high-frequency detail.} Same setup as Figure~\ref{fig:app_sem_knob_synthetic}. From top to bottom: \emph{peacock feather}, \emph{Mandelbrot fractal}, \emph{text paragraph}, and \emph{low-frequency noise}.}
    \label{fig:app_sem_knob_difficult}
\end{figure}

%% file: sections/appendix/knn_pooling.tex
\clearpage
\section{kNN Classification under Mean and Max Pooling}
\label{app:knn_pooling}

We measure the semantic informativeness of the latent tokens with a kNN classifier ($k{=}20$) on ImageNet-100, using the training set as gallery and the validation set as queries.
Since the number of active tokens varies per image, we pool the active register tokens of each image into a single vector.
In the main paper, we use mean pooling (\cref{fig:iso_knn_comparative}).
In this appendix, we additionally report max pooling.

The two poolings summarize the tokens differently.
Mean pooling averages over all active tokens, so every token contributes equally to the representation.
Tokens that carry semantic information add to the class signal, whereas tokens that carry little semantic information dilute it.
Max pooling instead retains, for each feature dimension, the largest activation across the active tokens.
A token thus affects the representation only in the dimensions where it exceeds all other tokens.
Max pooling is therefore less sensitive than mean pooling to the addition of tokens that carry little semantic information.
A higher accuracy under max pooling than under mean pooling hence suggests that some tokens dilute the semantic information carried by the others.

\begin{figure}[H]
    \centering
    \includegraphics[width=0.45\textwidth]{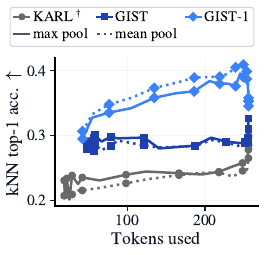}
    \caption{\textbf{The kNN comparison is robust to the choice of pooling.}
    kNN accuracy of \ourmethod, \karldagger, and \ourmethodone on the ImageNet-100 validation set, obtained by varying $\lrec$ for \karldagger and \ourmethod (fixing $\lsem = 0$), and by varying $\lsem$ for \ourmethodone (same setup as \cref{fig:iso_knn_comparative}).
    Accuracies are computed by max pooling (solid) and mean pooling (dotted) over the active tokens.
    The ordering of the three methods is the same under both poolings.
    For \ourmethodone, mean pooling yields a higher accuracy, and accuracy increases with the number of active tokens.
    For \ourmethod and \karldagger, max pooling yields a slightly higher accuracy, and accuracy remains approximately constant.}
    \label{fig:app_knn_pooling}
\end{figure}

In \cref{fig:app_knn_pooling}, we compare the kNN accuracy of \ourmethod, \ourmethodone, and \karldagger under mean and max pooling as we vary each method's target values (same setup as \cref{fig:iso_knn_comparative}).
We make two observations.
First, the ordering of the three methods is unchanged under max pooling.
\ourmethodone performs best, followed by \ourmethod and \karldagger, at any number of active tokens.
The conclusions of the main paper therefore do not depend on the choice of pooling.

Second, the relation between the two poolings differs across methods.
For \ourmethodone, mean pooling yields a higher accuracy than max pooling, and accuracy increases approximately linearly with the number of active tokens.
For \ourmethod and \karldagger, we observe the opposite.
Max pooling yields a slightly higher accuracy than mean pooling, and accuracy remains approximately constant as the number of active tokens increases.
This difference corresponds to the target varied for each method.
\begin{itemize}
    \item The curve of \ourmethodone is obtained by varying $\lsem$, whereas the curves of \ourmethod and \karldagger are obtained by varying $\lrec$ (with $\lsem = 0$ for \ourmethod).
    Lowering $\lsem$ demands tokens that preserve what the image depicts.
    The tokens added along this sweep therefore carry semantic information, and accuracy increases with their number.
    Under mean pooling, each of these tokens contributes to the representation, whereas under max pooling a token contributes only in the dimensions where it exceeds the others.
    Mean pooling can therefore benefit more from the added tokens, which is consistent with its higher accuracy for \ourmethodone.

    \item Lowering $\lrec$ instead demands tokens that reproduce the exact pixels of the image.
    The tokens added along this sweep encode pixel-level detail and carry little class information.
    Under mean pooling, these tokens dilute the semantic signal of the remaining tokens, whereas under max pooling they leave the largest activations largely unchanged.
    This is consistent with the slightly higher accuracy of max pooling for \ourmethod and \karldagger, and with their accuracy remaining approximately constant.
\end{itemize}

We note that these differences are subtle, and we do not draw strong conclusions from them.
They are, however, consistent with \cref{fig:iso_knn} and with the view that the targets control not only how many tokens an image requires, but also what those tokens encode.

%% file: sections/appendix/knn_k_sweep.tex
\clearpage
\section{kNN Classification for Different Values of $k$}
\label{app:knn_k_sweep}

\begin{figure}[H]
    \centering
    \includegraphics[width=0.75\textwidth]{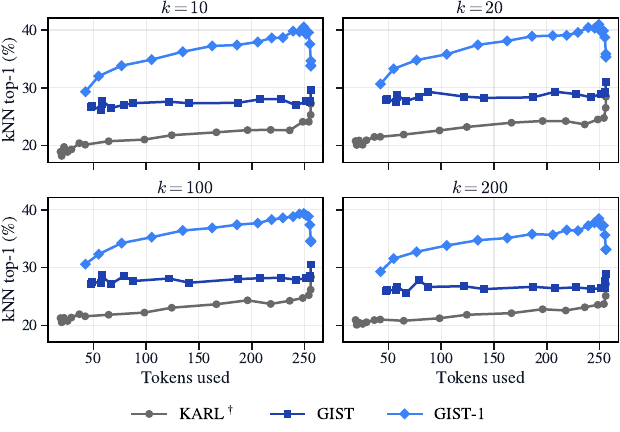}
    \caption{\textbf{The kNN comparison is robust to the choice of $k$.}
    kNN accuracy of \ourmethod, \karldagger, and \ourmethodone on the ImageNet-100 validation set for $k \in \{10, 20, 100, 200\}$ (same setup as \cref{fig:iso_knn_comparative}).
    The ordering of the three methods is the same for every $k$: \ourmethodone performs best, followed by \ourmethod and \karldagger, at any number of active tokens.
    Accuracy is highest at $k{=}20$ for all three methods and decreases slightly at $k{=}200$.}
    \label{fig:app_knn_k_sweep}
\end{figure}

In the main paper, we report kNN accuracy with $k{=}20$ neighbors (\cref{fig:iso_knn_comparative}).
In \cref{fig:app_knn_k_sweep}, we additionally report $k \in \{10, 100, 200\}$, with all other settings unchanged: mean pooling over the active tokens, the ImageNet-100 training set as gallery, and an unweighted majority vote over the retrieved neighbors.
The ordering of the three methods is the same for every $k$.
\ourmethodone performs best, followed by \ourmethod and \karldagger, at any number of active tokens, and its advantage over \karldagger lies between $6$ and $17$ points throughout.
The conclusions of the main paper therefore do not depend on the choice of $k$.

\Cref{tab:app_knn_k_sweep} reports the mean accuracy over the range of active tokens shared by all three methods, together with the advantage of \ourmethod and \ourmethodone over \karldagger.
The mean advantage varies little with $k$: it is $12.6$--$14.3$ points for \ourmethodone and $4.5$--$5.4$ points for \ourmethod.
Both advantages are largest at $k{=}10$ and decrease slightly as $k$ grows.

\begin{table}[H]
    \centering
    \caption{\textbf{Mean kNN accuracy of \karldagger, \ourmethod, and \ourmethodone for different values of $k$.}
    Mean kNN accuracy of \karldagger, \ourmethod, and \ourmethodone on the ImageNet-100 validation set for $k \in \{10, 20, 100, 200\}$ (same setup as \cref{fig:iso_knn_comparative}), averaged over the range of active tokens shared by all three methods ($47.7$--$256$ tokens).
    The last two columns report the advantage of \ourmethod and \ourmethodone over \karldagger in mean accuracy.
    For every $k$, \ourmethodone performs best, followed by \ourmethod and \karldagger.
    The advantage of \ourmethodone over \karldagger lies between $12.6$ and $14.3$ points, and that of \ourmethod over \karldagger between $4.5$ and $5.4$ points.
    Both advantages are largest at $k{=}10$ and decrease slightly as $k$ grows.}
    \label{tab:app_knn_k_sweep}
    \small
    \begin{tabular}{lccccc}
        \toprule
        & \multicolumn{3}{c}{Mean kNN top-1 (\%)} & \multicolumn{2}{c}{Advantage over \karldagger} \\
        \cmidrule(lr){2-4}\cmidrule(lr){5-6}
        $k$ & \karldagger & \ourmethod & \ourmethodone & \ourmethod & \ourmethodone \\
        \midrule
        $10$ & $22.0$ & $27.4$ & $36.3$ & $+5.4$ & $+14.3$ \\
        $20$ & $23.4$ & $28.6$ & $37.3$ & $+5.2$ & $+13.9$ \\
        $100$ & $23.3$ & $27.9$ & $36.3$ & $+4.6$ & $+13.0$ \\
        $200$ & $22.1$ & $26.6$ & $34.7$ & $+4.5$ & $+12.6$ \\
        \bottomrule
    \end{tabular}
\end{table}

%% file: sections/appendix/attentive_probing.tex
\clearpage
\section{Attentive Probing}
\label{app:attentive_probing}

To complement the parameter-free kNN evaluation in \cref{fig:iso_knn_comparative}, we probe the same frozen tokens with a trained classifier.
Specifically, for every target configuration of every tokenizer, we train a separate lightweight attentive probe, in which a learnable query cross-attends to the image's active tokens (i.e., with halted tokens masked out), and a linear classifier maps the resulting vector to class logits.
Architecture and training details of the probe are given in \cref{app:evaluation_protocol}.
We train the probe on the ImageNet-100 training set and report top-1 accuracy on the validation set for the same target values as in \cref{fig:iso_knn_comparative}.

\begin{figure}[H]
    \centering
    \begin{subfigure}{0.45\textwidth}
        \centering
        \includegraphics[width=\textwidth]{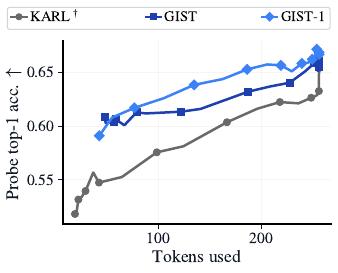}
        \caption{\ourmethod vs. \karldagger vs. \ourmethodone}
        \label{fig:iso_probe_comparative}
    \end{subfigure}
    \par\medskip
    \begin{subfigure}{\textwidth}
        \centering
        \includegraphics[width=\textwidth]{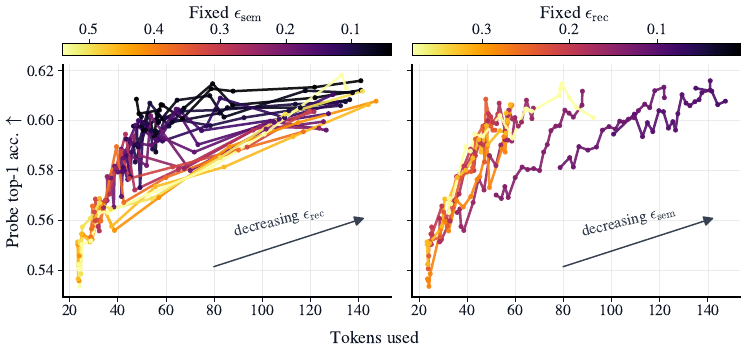}
        \caption{Varying one target at a time}
        \label{fig:iso_probe}
    \end{subfigure}
    \caption{\textbf{The advantage of semantic control holds under a trained attentive probe.}
    Both panels report the best validation top-1 accuracy over the 10 training epochs for each target configuration.
    (a)~Attentive-probe accuracy of \ourmethod, \karldagger, and \ourmethodone as we vary each method's target values (same setup as \cref{fig:iso_knn_comparative}).
    The ordering from the kNN evaluation is preserved, with narrower but consistent gaps: \ourmethodone and \ourmethod outperform \karldagger at every token count.
    (b)~Attentive-probe accuracy trajectories of \ourmethod as we vary one target with the other held fixed (shown by color).
    Lowering $\lsem$ increases probe accuracy at a much steeper rate than lowering $\lrec$.
    Together, the probe shows that conditioning on $\lsem$ not only exposes semantic information more readily, but also makes the tokens carry more of it.
    }
    \label{fig:token_space_isocurves_probe}
\end{figure}

In \cref{fig:iso_probe_comparative}, we compare the attentive-probe accuracy of \ourmethod, \ourmethodone, and \karldagger as we vary each method's target values.
The ordering from the kNN evaluation is preserved (\ourmethodone performs best, followed by \ourmethod and \karldagger), but the gaps narrow.
\ourmethodone and \ourmethod lead \karldagger by $+6$ and $+5$ points at $64$ tokens and by $+4$ and $+2$ points at $192$ tokens, compared to $+12$ and $+6$ points at $64$ tokens and $+15$ and $+4$ points at $192$ tokens under kNN (\cref{fig:iso_knn_comparative}).
We draw two conclusions.
First, the narrower gaps show that a considerable part of the kNN advantage lies in how \textit{readily} semantic information can be read out.
A kNN classifier only sees the class information that mean pooling exposes directly, whereas a trained probe can attend to the relevant tokens and combine them.
Second, the gaps do not close.
In other words, even a probe that is free to select and weight tokens recovers more class-relevant information from \ourmethod and \ourmethodone tokens than from \karldagger tokens.
Semantic control therefore not only makes semantic information more accessible, but also increases how much of it the tokens carry.

In \cref{fig:iso_probe}, we repeat the single-target trajectories of \cref{fig:iso_knn} with probe accuracy in place of kNN accuracy.
The picture mirrors the main paper.
Lowering $\lrec$ (left) yields long, shallow curves along which accuracy grows only slowly with the number of tokens, whereas lowering $\lsem$ (right) yields short, steep curves that gain accuracy over only a few added tokens.
Lowering the semantic target is thus the more token-efficient way to make tokens semantically informative, whether the readout is a nearest-neighbor lookup or a trained probe.

\cref{tab:probe_best_vs_final} compares best-epoch and final-epoch accuracies at $64$ and $192$ tokens.
The ordering is unchanged, and the advantages of \ourmethod and \ourmethodone over \karldagger change by at most $0.8$ percentage points.

\begin{table}[H]
    \caption{\textbf{Best-epoch vs.\ final-epoch attentive-probe accuracy.}
    Top-1 accuracy (\%) at $64$ and $192$ tokens, obtained by linear interpolation between neighboring target configurations on each method's curve.
    For each configuration, \emph{Best epoch} uses the highest validation accuracy over the 10 training epochs, as in \cref{fig:iso_probe_comparative}, while \emph{Final epoch} uses the validation accuracy after epoch 10.}
    \label{tab:probe_best_vs_final}
    \centering
    \begin{tabular}{@{}llcc@{}}
    \toprule
    Method & Tokens & Best epoch & Final epoch \\
    \midrule
    \karldagger & 64 & 55.2 & 54.8 \\
    \ourmethod & 64 & 60.3 & 59.9 \\
    \ourmethodone & 64 & 61.2 & 60.5 \\
    \midrule
    \karldagger & 192 & 61.4 & 60.6 \\
    \ourmethod & 192 & 63.3 & 63.1 \\
    \ourmethodone & 192 & 65.4 & 65.4 \\
    \bottomrule
    \end{tabular}
\end{table}